\documentclass[journal]{IEEEtran}
\usepackage{cite}

\usepackage[pdftex]{graphicx}
\graphicspath{{imgs/}}
\ifCLASSINFOpdf
\else
\fi
\usepackage{amsmath}
\usepackage{amssymb}
\usepackage[linesnumbered, ruled, vlined]{algorithm2e}
\usepackage[caption=false,font=normalsize,labelfont=sf,textfont=sf]{subfig}

\usepackage{stfloats}
\begin{document}
%
\title{Neural Architecture Search using Covariance Matrix Adaptation Evolution Strategy}
%
%
%

\newcommand{\comm}[1]{}

\author{Nilotpal~Sinha,
	Kuan-Wen~Chen,~\IEEEmembership{Member,~IEEE}
	\comm{and~Jane~Doe,~\IEEEmembership{Member,~IEEE}
		
		\thanks{M. Shell was with the Department
			of Electrical and Computer Engineering, Georgia Institute of Technology, Atlanta,
			GA, 30332 USA e-mail: (see http://www.michaelshell.org/contact.html).}
		\thanks{J. Doe and J. Doe are with Anonymous University.}
		\thanks{Manuscript received April 19, 2005; revised August 26, 2015.}}
}

%
%

\markboth{Under Review}%
{Shell \MakeLowercase{Sinha \textit{et al.}}:}
%



\maketitle

\begin{abstract}
Evolution-based neural architecture search requires high computational
resources, resulting in long search time. In this work, we propose a framework
of applying the Covariance Matrix Adaptation Evolution Strategy (CMA-ES) to the
neural architecture search problem called CMANAS, which achieves better results
than previous evolution-based methods while reducing the search time
significantly. The architectures are modelled using a normal distribution, which
is updated using CMA-ES based on the fitness of the sampled population. 
We used the accuracy of a trained one shot model (\textit{OSM}) on the validation
data as a prediction of the fitness of an individual architecture to reduce the
search time. We also used an architecture-fitness table (AF table) for keeping
record of the already evaluated architecture, thus further reducing the search
time. CMANAS finished the architecture search on CIFAR-10 with the top-1 test
accuracy of 97.44\% in 0.45 GPU day and on CIFAR-100 with the top-1 test accuracy
of 83.24\% for 0.6 GPU day on a single GPU. The top architectures from the
searches on CIFAR-10 and CIFAR-100 were then transferred to ImageNet, achieving
the top-5 accuracy of 92.6\% and 92.1\%, respectively.

\end{abstract}


%
\IEEEpeerreviewmaketitle

\section{Introduction}
%
%
%
%
\comm{\IEEEPARstart{T}{his} demo file is intended to serve as a ``starter
	 file'' for IEEE journal papers produced under \LaTeX\ using
	  IEEEtran.cls version 1.8b and later.
I wish you the best of success.}

\begin{figure}[t]
	\centering
	\begin{center}
		\includegraphics[width=0.95\linewidth]{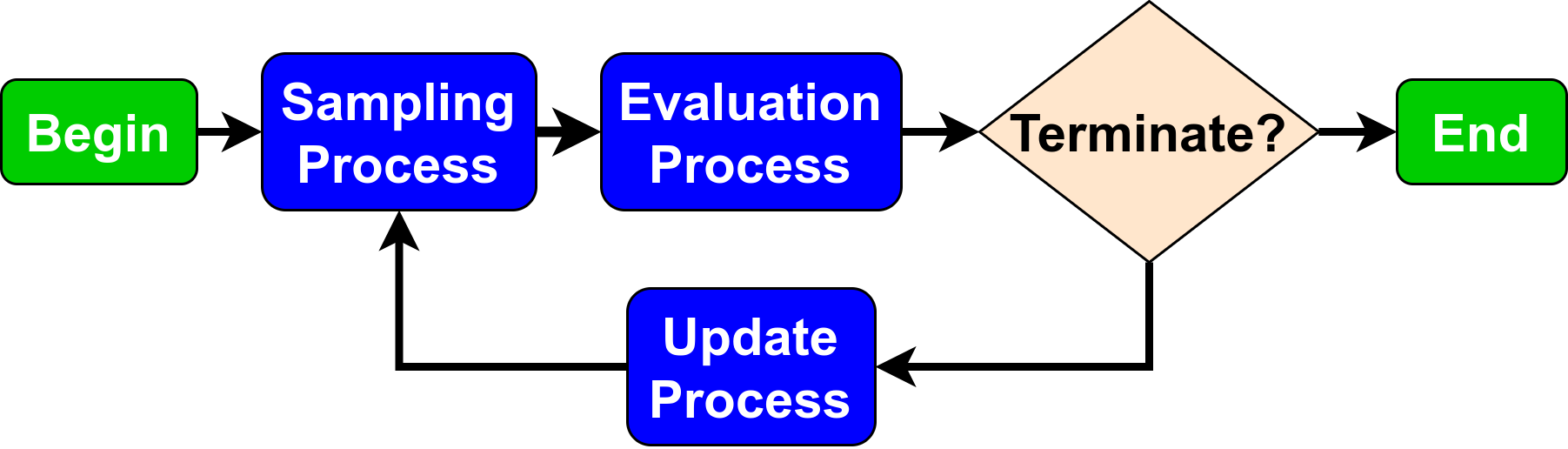}
	\end{center}
	\caption{Abstract illustration of Neural Architecture Search methods.}
	\label{fig:NAS_algorithms}
\end{figure}

\IEEEPARstart{I}{n} the recent years, convolutional neural networks
(CNNs) have been very instrumental in solving various computer vision
problems. However, the CNN architectures (such as 
AlexNet\cite{krizhevsky2012imagenet}, ResNet\cite{he2016deep}, 
DenseNet\cite{huang2017densely}, VGGNet\cite{simonyan2014very})
have been designed mainly by humans, relying on their intuition and
understanding of specific problem. \textit{Neural architecture search} (NAS)
tries to replace the reliance on human intuition with an automated search of the
neural architecture. Recent NAS methods
\cite{elsken2018neural}\cite{zoph2016neural}\cite{pmlr-v80-pham18a}
have shown promising results in the field of computer vision but most
of these methods consume a huge amount of computational power. Any NAS
method has three parts (Fig.~\ref{fig:NAS_algorithms}): \textit{sampling}
process, \textit{evaluation} process and \textit{update} process. The
epoch in the NAS algorithm begins with sampling architecture from a given
search space (i.e. sampling process), which is then sent to the evaluation
process for the performance evaluation. On the basis of the evaluation metric,
the NAS algorithm will update the sampling process (i.e. update
process) in order to sample better performing architecture in the next
epoch. The algorithm stops upon satisfying a stopping criterion.

Evolutionary algorithm (EA)-based NAS samples a population of
architectures during the sampling process, which is updated during the
update process on the basis of performance of the architectures from the
evaluation process. Reinforcement learning (RL)-based NAS has a RL
agent sampling architecture during the sampling process, which is updated
during the update process depending on the performance of the architecture
determined by the evaluation process. Both types of methods require huge
computational resources, resulting in long search time. For example,
the method proposed in \cite{real2019regularized} required 3150 GPU days of
evolution and that discussed in \cite{zoph2018learning} required 1800 GPU days of
RL search. This is attributed to the evaluation process wherein each architecture
is trained from scratch for a certain number of epochs in order to evaluate
its performance on the validation data. Recently proposed gradient-based methods
such as 
\cite{liu2018darts2}\cite{dong2019searching}\cite{xie2018snas}\cite{dong2019one}
\cite{chen2019progressive} have reduced the search time
by sharing weights among the architectures. However, these gradient-based
methods highly depend on the given search space and suffer from
premature convergence to the local optimum as shown in
\cite{chen2019progressive}\cite{Zela2020Understanding}.

\begin{figure*}[t]
	\centering
	\begin{center}
		\includegraphics[width=0.9\linewidth]{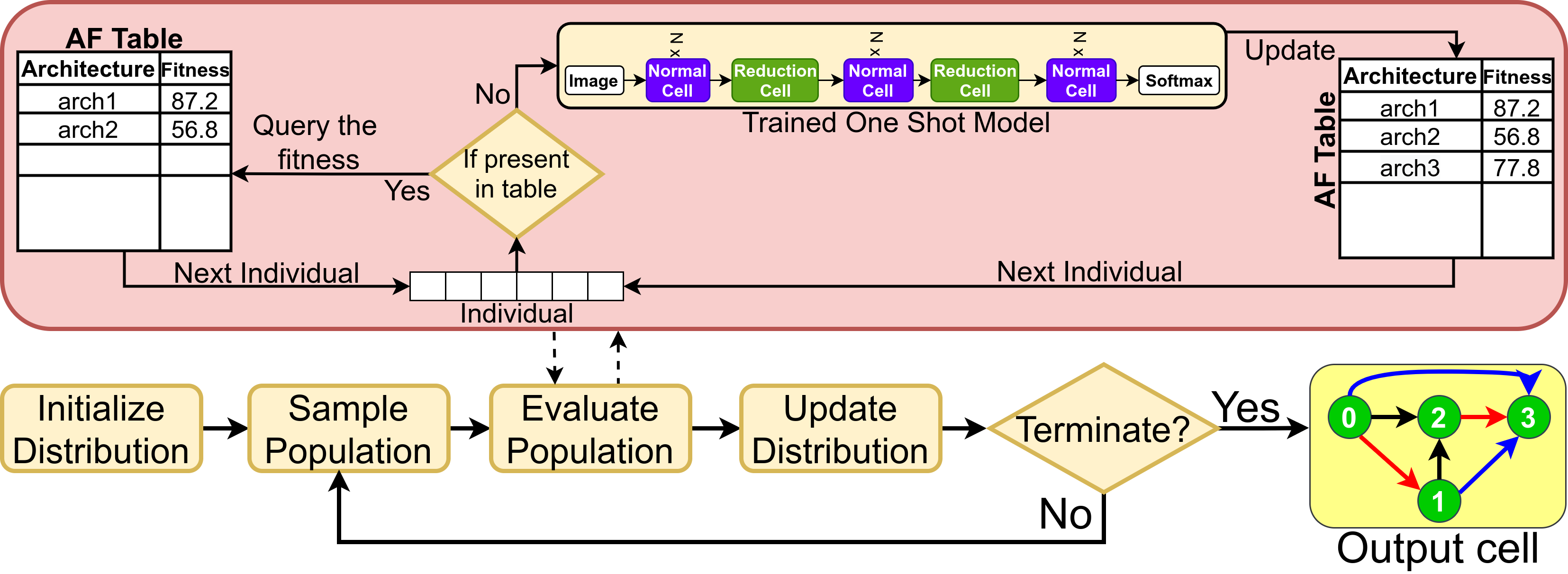}
	\end{center}
	\caption{Illustration of the general framework of CMANAS.}
	\label{fig:arch_search}
\end{figure*}

In this work, we propose a method called CMANAS (\textit{Neural
Architecture using Covariance Matrix Adaptation Evolution Strategy}) which
is summarized in Fig.~\ref{fig:arch_search}. Here, the neural architecture is
represented by a \textit{normal distribution}. In every epoch, the distribution is
first used to sample a population of architectures (i.e. sampling process) and
then the distribution is updated using the covariance matrix adaptation evolution
strategy (CMA-ES) \cite{hansen2001completely} (i.e. update process) on the basis
of the performance of the population of architectures (i.e. evaluation process).
We used a trained \textit{one shot model} (OSM) to evaluate each architecture
instead of training each architecture from scratch. This resulted in reduced
search time because of the weight sharing among all the architectures in the one
shot model. We also used an architecture-fitness table (AF table) to maintain a
record of the already evaluated architectures, which further resulted in reducing
the search time.

Our contributions could be summarized as follows:
\begin{itemize}
	\item We designed a framework of applying the covariance matrix adaptation
	evolution strategy to the NAS problem with reduced computational
	requirements by using a trained one shot model for evaluating the
	performance/\textit{fitness} of an architecture. The architectures are
	modelled by a multivariate normal distribution, which is updated using
	CMA-ES depending on the fitness estimated by the trained OSM.
	
	\item We used an architecture-fitness table (AF table) for maintaining the
	records of the already evaluated architectures in order to skip the process of
	re-evaluating an already evaluated architecture and thus reducing the search
	time.
	
	\item We applied our method to two different search spaces to show that
	CMANAS is search space independent (i.e. search space agnostic) as is not the
	case for the gradient-based NAS methods.
	
	\item We also created a visualization of the architecture search
	performed by CMANAS to get insights into the search process. We found that
	the first phase of the search is predominantly an \textit{exploration} phase
	wherein CMANAS explores the given search space. This is followed by an
	\textit{exploitation} phase (i.e. convergence to an architecture).
\end{itemize}

\section{Related Work}
Searching the neural architecture automatically by using an algorithm (i.e. NAS)
is an alternative to the architectures designed by humans, and in the recent years,
these NAS methods have attracted increasing interest because of its promise of
an automatic and efficient search of architectures specific to a task. Early NAS
approaches \cite{stanley2002evolving}\cite{stanley2009hypercube} optimized both the
neural architectures and the weights of the network using evolution. However,
their usage was limited to shallow networks. Recent NAS methods
\cite{zoph2016neural}\cite{pmlr-v80-pham18a}\cite{real2019regularized}\cite{zoph2018learning}\cite{real2017large}\cite{liu2018hierarchical}\cite{xie2017genetic}
perform the architecture search separately while using gradient descent for
optimizing the weights of the architecture for its evaluation. This has made
possible the search of deep networks. The various NAS methods can be classified
into two categories on the basis of the different methods used in the update
process in Fig.~\ref{fig:NAS_algorithms}. These are \textit{gradient-based}
methods and \textit{non-gradient based} methods.

\textbf{Gradient-Based Methods:} These methods begin with a random neural
architecture, which can be regarded as the sampling process in 
Fig.~\ref{fig:NAS_algorithms}. The neural architecture is then updated using
the gradient information on the basis of its performance on the validation data
(i.e. update process in Fig.~\ref{fig:NAS_algorithms}). In general, these methods
\cite{liu2018darts2}\cite{dong2019searching}\cite{xie2018snas}\cite{dong2019one}
relax the discrete architecture search space to a continuous search space by
using an one shot model (OSM). The performance of the OSM on the validation data is
used for updating the architecture using gradients. As the OSM shares weights
among all architectures in the search space, these methods take lesser time in the
evaluation process in Fig.~\ref{fig:NAS_algorithms} and thus shorter search time.
However, these methods suffer from the overfitting problem wherein the
resultant architecture shows good performance on the validation data but exhibits
poor performance on the test data. This can be attributed to its preference for
parameter-less operations in the search space, as it leads to rapid gradient
descent\cite{chen2019progressive}. Some regularization techniques have been
introduced to tackle this problem, such as early stopping
\cite{Zela2020Understanding}, search space regularization
\cite{chen2019progressive} and architecture refinement
\cite{chen2019progressive}. In contrast to these gradient-based methods, our
method does not suffer from the overfitting problem because of its stochastic
nature.

\textbf{Non-Gradient Based Methods:} \comm{Here, no gradient information is used
in the update process in Fig.~\ref{fig:NAS_algorithms}.} These methods include
reinforcement learning (RL) methods and evolutionary algorithm (EA) methods.
In the RL methods, an agent is used for the generating neural architecture
(i.e. sampling process in Fig.~\ref{fig:NAS_algorithms}). The agent is then
trained (i.e. update process in Fig.~\ref{fig:NAS_algorithms}) to generate
architectures in order to maximize its expected accuracy on the validation data
(calculated in the evaluation process in Fig.~\ref{fig:NAS_algorithms}). 
In \cite{zoph2016neural}\cite{zoph2018learning}, a recurrent neural network (RNN)
is used as an agent for sampling the neural architectures. These sampled
architectures are then trained from scratch to convergence in order to get their
accuracies on the validation data (i.e. evaluation process in
Fig.~\ref{fig:NAS_algorithms}). These accuracies are then used for updating
the weights of the RNN agent by using policy gradient methods. Because of the
huge computational requirement of training the architectures from scratch in the
evaluation process, both of these methods suffered from long search time. This was
improved in \cite{pmlr-v80-pham18a} by using a single directed acyclic graph (DAG)
for sharing the weights among all the sampled architectures, thus resulting
in reduced computational resources.

The EA based NAS methods begin with a population of architectures which can be
regarded as the sampling process in Fig.~\ref{fig:NAS_algorithms}. Each
architecture in the population is evaluated on the basis of its performance on the
validation data (evaluation process in Fig.~\ref{fig:NAS_algorithms}). The
popluation is then evolved (i.e. update process in Fig.~\ref{fig:NAS_algorithms})
on the basis of the performance of the population. Methods such as those proposed
in \cite{real2019regularized}\cite{xie2017genetic} used gradient descent for
optimizing the weights of each architecture in the population from scratch in
order to determine their accuracies on the validation data as their fitness during
the evaluation process, resulting in huge computational requirements. In order to
speed up the training process, in \cite{real2017large}, the authors introduced
weight inheritance wherein the architectures in the post-update process population
inherit the weights of the pre-update process population, resulting in bypassing
the training from scratch. However, the speed up gained is less as it still needs
to optimize the weights of the architecture. Methods such as that proposed in
\cite{sun2019surrogate} use a random forest for predicting the performance of the
architecture during the evaluation process, resulting in a high speed up as
compared to previous EA methods. However, its performance was far from the
state-of-the-art results. In contrast, our method achieved better results than
previous EA methods while using significantly less computational resources. CMA-ES
has shown good performance in many high-dimensional continuous optimization
problems such as fine-tuning the hyperparameters of the
CNN \cite{loshchilov2016cma}. However, to the best of our knowledge,
CMA-ES has not been applied to the NAS problem because of the discrete nature of
the problem.

\comm{FairNAS \cite{chu2019fairnas}, NSGANetV2 \cite{lu2020nsganetv2} has
also proposed evolutionary search with weight sharing which has two steps for searching neural
architecture where they optimizes the supernet in the first step and then FairNAS performs
architecture search using evolutionary method with the trained supernet as the evaluator in the second
step while NSGANetV2 uses the weights from trained supernet to initialize weights of an architecture
and train it using gradient descent for some epochs to evalauates its fitness during the architecture
search. In contrast, our method combines both the training and search process in one single stage.
FairNAS and NSGANetV2 solves the search problem as a multi-objective problem whereas our
	method solves it as a single objective problem.}

\comm{
\hfill mds
 
\hfill August 26, 2015
}

\section{Proposed Approach}
\label{methods}

\begin{figure}[t]
	\centering
	\subfloat[]{
		\label{subfig:macro_arch_a}
		\includegraphics[width=0.25\linewidth,scale=0.6]{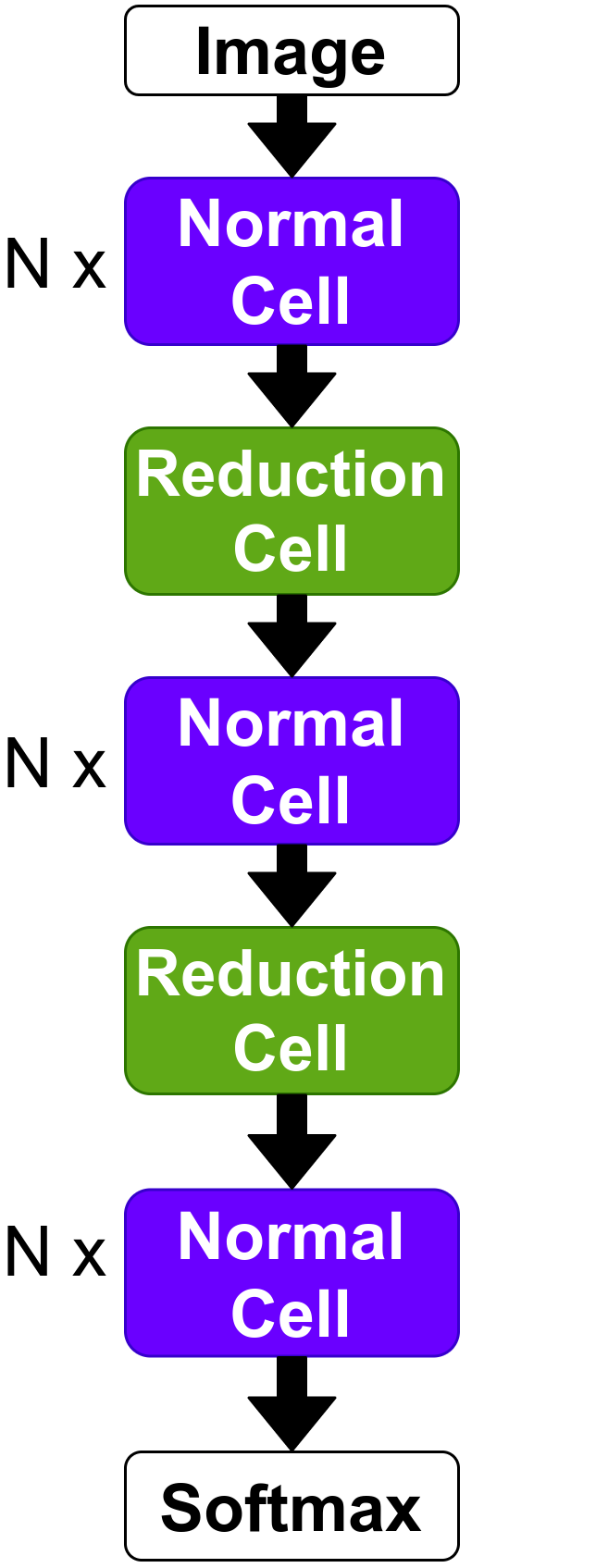}}
	\qquad
	\subfloat[]{
		\label{subfig:macro_arch_b}
		\includegraphics[width=0.53\linewidth,scale=1]{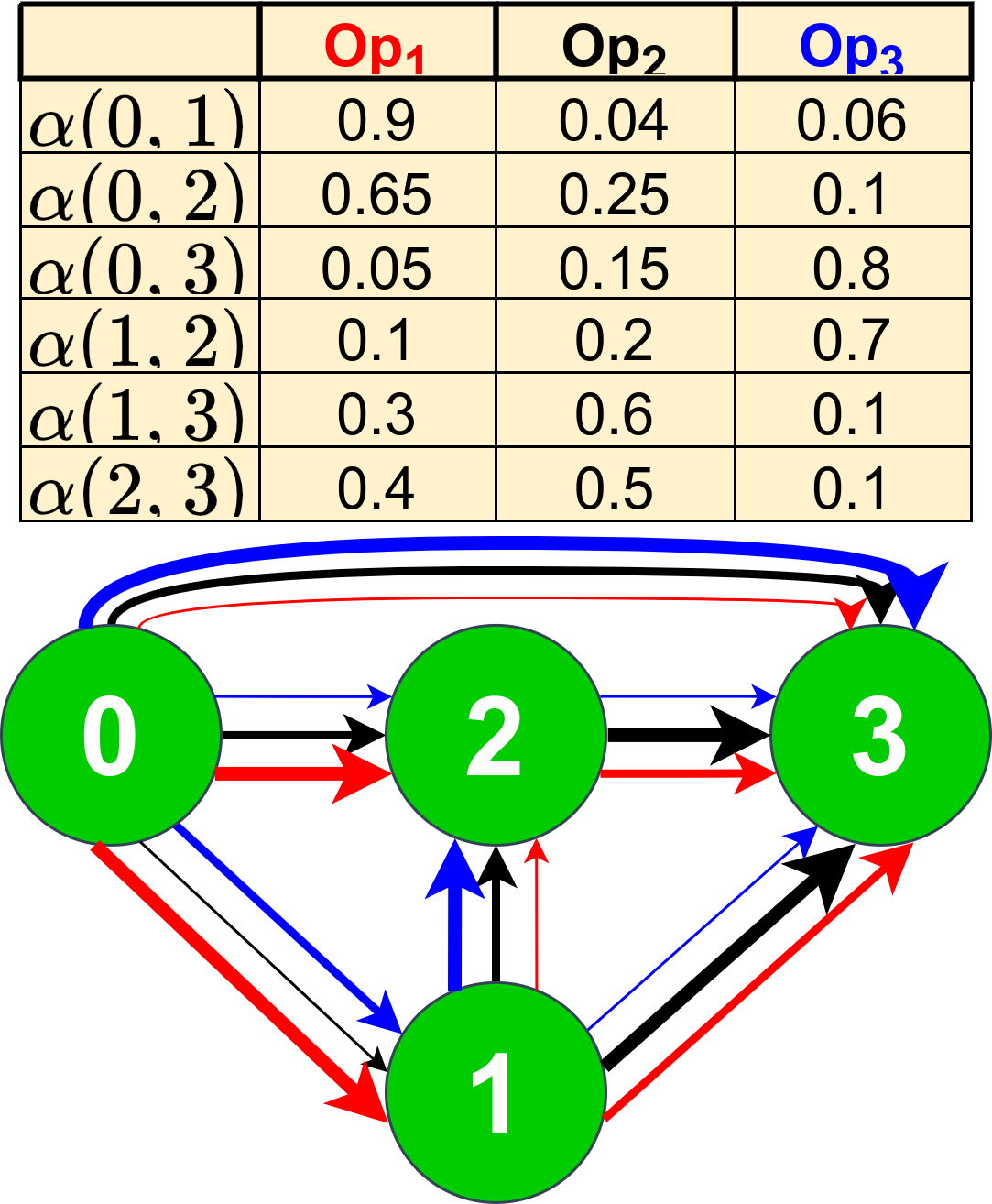}}
	\caption{(a) Architecture created by stacking cells.
		(b) Architecture representation ($\alpha$) of a cell in the one
		shot model with three different operations $Op(.)$ in the
		search space $\mathcal O$. The colors of the arrows between any two nodes represent the different operations and the thickness of
		the arrow is proportional to the weight of the corresponding operation.}
	\label{fig:macro_arch}
\end{figure}

\subsection{Search Space}
\label{subsect:search_space}
The choice of the search space can affect the quality of the searched
architecture. CMANAS searches for both operations and connections which is
in contrast to previous EA-based NAS
\cite{xie2017genetic}\cite{sun2019surrogate}\cite{elsken2018efficient}
\cite{suganuma2017genetic}, focusing on one facet of the architecture search,
e.g. connections and/or hyper-parameters. This makes our search space more
comprehensive. The success of the recent hand-crafted CNN architectures is
attributed to their sharing similar characteristics of repeating motifs
\cite{he2016deep}\cite{huang2017densely}\cite{szegedy2016rethinking}.
Therefore, in \cite{zoph2018learning}\cite{zhong2018practical}, the researchers
proposed to search for such motifs called cells, instead of the whole
architecture. In this work, we used this cell-based search space, which has been
successfully employed in recent works
\cite{pmlr-v80-pham18a}\cite{real2019regularized}\cite{zoph2018learning}\cite{liu2018darts2}\cite{dong2019searching}\cite{dong2019one}\cite{lu2020multi}\cite{liu2018progressive} and allows us to be consistent with these works. As illustrated in
Fig \ref{fig:macro_arch}(a), the architecture is created
by staking together cells of two types: \textit{normal} cells which
preserve the dimentionality of the input with a stride of one and
\textit{reduction} cells which reduce the spatial dimension with a stride
of two.

To construct both types of cells, we used \textit{directed acyclic graphs}
(DAGs) containing \textit{n} nodes and edge $(i, j)$ between any two nodes
representing an operation from the search space with $N_{ops}$
different operations. In this work, we applied our method to two different
search spaces: \textit{Search space 1 (\textbf{S1})}\cite{liu2018darts2} and
\textit{Search space 2 (\textbf{S2})}\cite{Dong2020NAS-Bench-201}. In
\textit{S1}, we search for both normal and reduction cells where each node
$x^{(j)}$ maps two inputs to one output. The two inputs for $x^{(j)}$ in
cell $k$ are picked from the outputs from previous nodes $x^{(i)}$ in 
cell $k$ (i.e. $i < j$), output from previous cell $c_{k-1}$ and output from the
previous-to-previous cell $c_{k-2}$. In \textit{S2}, we search for only normal
cells, where each node $x^{(j)}$ is connected to the previous node $x^{(i)}$
(i.e. $i < j$).

\subsection{Representation of Architecture}
\label{subsect:representation}
\begin{figure}[t]
	\centering
	\begin{center}
		\includegraphics[width=0.9\linewidth]{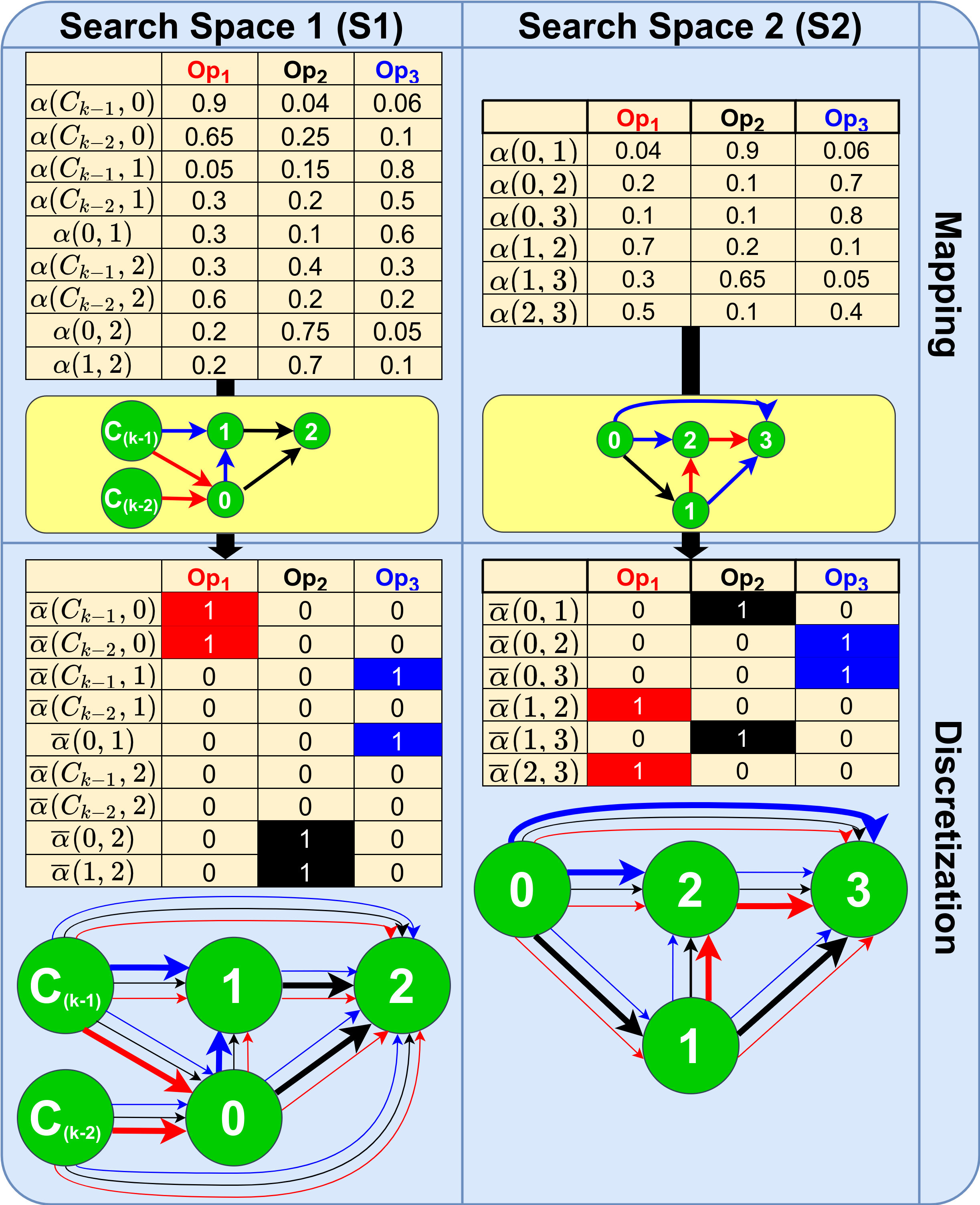}
	\end{center}
	\caption{Process of evaluating architecture using
		the trained one shot model with three nodes for S1 and four
		nodes for S2. \textit{Top}: mapping process wherein the
		architecture parameter $\alpha$	is mapped to its corresponding
		architecture. $c_{k-1}$ and $c_{k-2}$ refer to outputs from the
		previous cell and previous-to-previous cell respectively.
		\textit{Bottom}: discretization process wherein
		$\hat\alpha$ is created from the derived architecture and is
		copied to the ONS. The colors of the arrows between any two nodes
		represent the different operations and the thickness of
		the arrow is proportional to the weight of the corresponding operation.}
	\label{fig:evalute}
\end{figure}
Evaluating an architecture involves training it from scratch for some
epochs and then evaluating it on the basis of its performance on validation data,
leading to a long search time\cite{real2019regularized}\cite{zoph2018learning}.
Instead, we used a \textit{one shot model} (\textit{\textbf{OSM}})
\cite{liu2018darts2}, which shares the weights among all architectures by
treating all the architectures as the subgraphs of a supergraph. As illustrated in
Fig \ref{fig:macro_arch}(b), a cell in the one shot model is represented
by an \textit{architecture parameter}, $\alpha$. It represents the
weights of different operations $Op(.)$ in the given operation space
$\mathcal O$ (i.e. the search space of NAS) between a pair of nodes where
each $Op(.)$ represents some funtion to be applied to some node
$x^{(i)}$. In OSM, the directed edge from node $i$ to node $j$ is the
weighted sum of all $Op(.)s$ in $\mathcal O$ where the $Op(.)s$ are
weighted by normalized $\alpha^{(i,j)}$ using softmax. This can be
written as:
\begin{equation}
	\label{eq:ons}
	f^{(i,j)}(x^{(i)}) = \sum_{Op \in \mathcal O } 
	\frac{exp(\alpha^{(i,j)}_{Op})}
	{\sum_{Op' \in \mathcal O } exp(\alpha^{(i,j)}_{Op'})} Op(x^{(i)})
\end{equation}
where $\alpha^{(i,j)}_{op}$ represents the weight of the operation $Op(.)$ in
the operation space $\mathcal O$ between node $i$ and node $j$.
Each $\alpha$ for a normal cell and a reduction cell is represented by a
matrix (Fig. \ref{fig:macro_arch}(b)) with columns representing the
weights of different operations $Op(.)s$ from the operation space $\mathcal
O$ and rows representing the edge between two nodes. This design choice
allows us to skip the individual architecture training from scratch for
its evaluation because of the weight-sharing nature of OSM, thus resulting
in a significant reduction of search time. As discussed in
Section~\ref{subsect:search_space}, an architecture is derived from
$\alpha$ for the two search spaces through a \textit{mapping} process
in the following ways:

\begin{itemize}
	\item \textit{Search space 1 (S1)}: For each node, the top two distinct
	input nodes are chosen from all previous nodes on the basis of the weights
	of all the operations in the search space.
	\item \textit{Search space 2 (S2)}: For each edge between any two
	nodes, the top operation is chosen on the basis of weights of all the
	operations in the search space.
\end{itemize}
Fig.~\ref{fig:evalute} illustrates the mapping process with three
operations in both S1 and S2 and three nodes in S1 and four nodes in S2.

\subsection{Performance Estimation}
\label{subsect:performance}
\begin{figure}[t]
	\centering
	\begin{center}
		\includegraphics[width=0.98\linewidth]{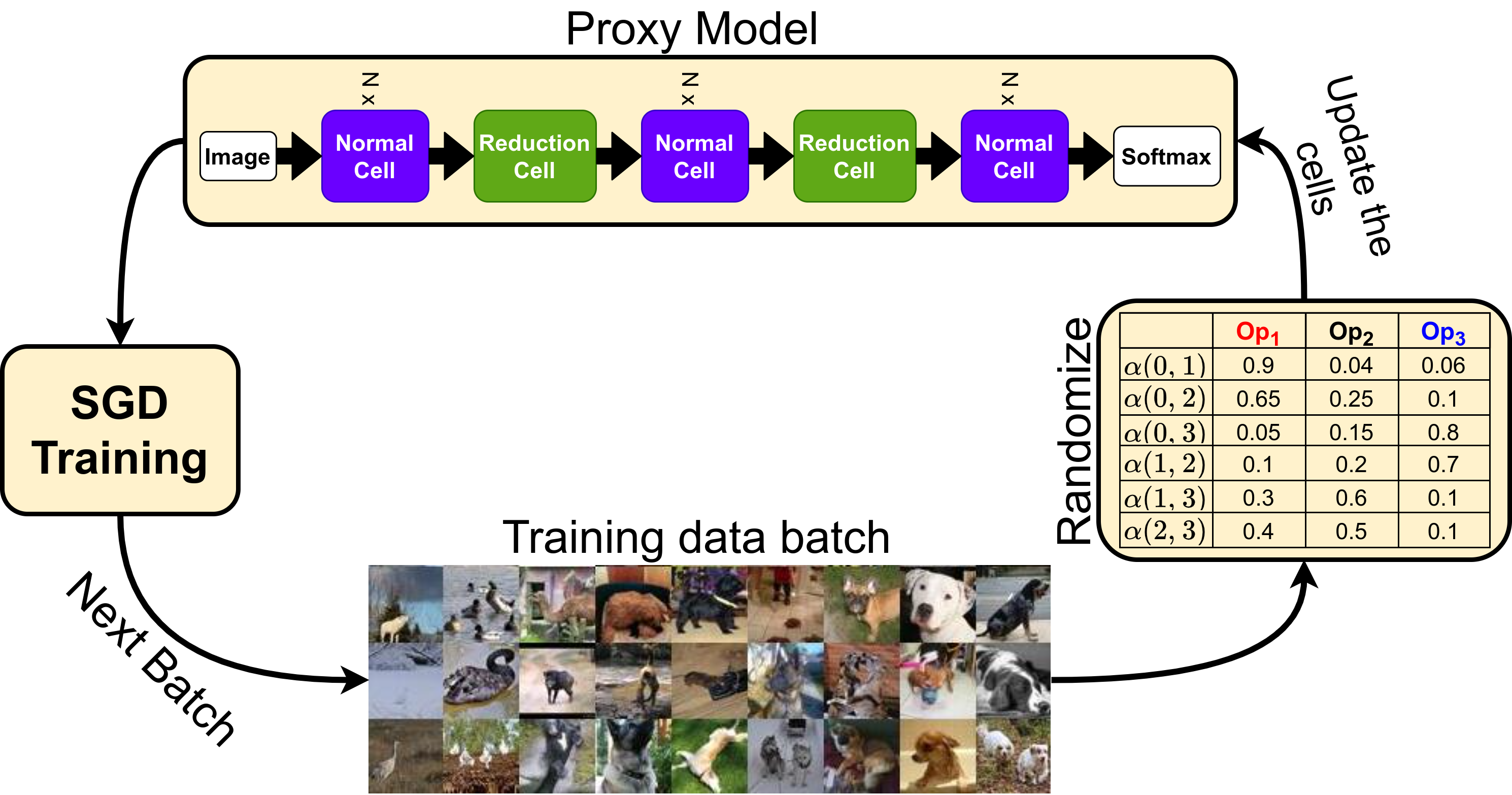}
	\end{center}
	\caption{Abstract illustration of training the one shot model in an
		epoch.}
	\label{fig:training}
\end{figure}

The performance of an architecture is calculated by the trained one shot
model using the validation data. The OSM is trained for a certain number
of epochs using \textit{Stochastic Gradient Descent}
(SGD\cite{sutskever2013importance}) with momentum. The training of the OSM
begins with randomly initializing the weights of the OSM. Then for each
training batch in an epoch, the architecture parameter, $\alpha$ of the
OSM is updated to random values (as shown in Fig. \ref{fig:training}) so
that no particular sub-graph (i.e. architecture) receives most of the
gradient updates of the super-graph (i.e. OSM). The algorithm is summarized in
Algorithm~\ref{algo:training_OSM} and its implementation details discussed in
Section~\ref{subsubsect:training_settings}.

The trained OSM from Algorithm~\ref{algo:training_OSM} is then used to
evaluate an architecture on the basis of its accuracy on the validation data,
also known as the \textit{fitness} of the architecture. The process of
evaluation follows two steps sequentially (as illustrated in Fig.
\ref{fig:evalute}):
\begin{itemize}
	\item \textit{Mapping} process: Here, the architecture, $\mathcal A$,
	is derived from the architecture parameter $\alpha$, on the basis of the
	search space used (Section~\ref{subsect:representation}).
	\item \textit{Discretization} process: The derived architecture is
	then used to create a new architecture parameter called
	\textit{discrete architecture} parameter, $\bar\alpha$ with the
	following entries:
	\begin{equation}
		\bar{\alpha}^{(i,j)}_{Op} = 
		\begin{cases}
			1, \text{if $Op(x^{(i)})$ present in $\mathcal A$}\\
			0,  \text{otherwise}\\
		\end{cases}    
	\end{equation}
	 $\bar\alpha$ is then copied to the OSM for evaluating its accuracy
	 on the validation data, which is the \textit{fitness} of the
	 architecture $\mathcal A$.
\end{itemize}

\begin{algorithm}[t]
	\comm{https://arxiv.org/pdf/1903.03614.pdf}	
	\caption{Training one shot model}	
	\label{algo:training_OSM}
	\SetAlgoLined
	\KwIn{One shot model $M$, architecture parameter \textit{$\alpha$},
		training data $\mathfrak{D}_{tr}$, total epochs $N_{epochs}$, initial learning rate $\eta_{max}$, minimum learning rate $\eta_{min}$,
		weight decay \textit{$\lambda$}, momentum $\rho$.}
	\KwOut{Trained one shot model $M$.}
	$W \gets$ Ramdomly initialize weights of M\;
	$\Delta v \gets 0$ (Initialize the momentum term)\;
	\For{ $\tau \gets 1$ to $N_{epochs}$ }{
		$\eta_{\tau} \gets \eta_{\tau} +\frac{1}{2}(\eta_{max}-\eta_{min})(1+\cos(\frac{\tau}{N_{epochs}}\pi))$\;
		\For{each batch ($\mathfrak{B}$) in $\mathfrak{D}_{tr}$} {
			$\alpha \gets$ Update to random values\;
			$\hat W \gets W - \eta_{\tau}\cdot\rho\cdot\Delta v$\;
			$\mathcal{L}(\hat W, \mathfrak{B})\gets$ Cross-entropy on the
			batch\;
			$\Delta v \gets \rho\cdot\Delta v + (1-\rho)\cdot\nabla_{W}\mathcal{L}(\hat W, \mathfrak{B})$\;
			$W \gets (1-\lambda)W - \eta_{\tau}\Delta v$\;
		}
	}		
\end{algorithm}

\subsection{CMA-ES}
\begin{algorithm}[t]
	\caption{CMANAS}	
	\label{algo:CMANAS}
	\SetAlgoLined
	\KwIn{Trained one shot model $M$, validation data
		$\mathfrak{D}_{va}$, total generations $N_{gen}$,
		population size $N_{pop}$.}
	\KwOut{Searched architecture, $\textbf{m}^{(N_{gen})}$.}
	$g \gets 0$ (Initialize the epoch counter)\;
	
	Initialize CMA-ES with $\textbf{m}^{(g)} \gets \vec{0}$,
	$\textbf{C}^{(g)} \gets \textbf{I}$ and $\sigma^{(g)}$\;
	
	\While{ $g \leq N_{gen}$ }{
		Sample population, $\mathcal{P}$, of $N_{pop}$ individuals using Eq. \ref{eq:sampling}\;
		
		\For{each individual architecture ($\mathcal{A}$) in
			$\mathcal{P}$} {
			\uIf{$\mathcal{A}$ already present in AF table}{
				$fitness(\mathcal{A}) \gets$ entry in the AF table\;
			}
			\Else{
				$fitness(\mathcal{A}) \gets$ Evalate $\mathcal{A}$ using $M$\;
				Insert $fitness(\mathcal{A})$ to the AF table\;
			}
		}
		Update $\textbf{m}^{(g)}$, $\sigma^{(g)}$ and $\textbf{C}^{(g)}$
		using Eq.~\ref{eq:mean_update}, Eq.~\ref{eq:sigma_update},
		Eq.~\ref{eq:cov_update} in CMA-ES\;
		$g \gets g + 1$\;
	}		
\end{algorithm}
Covariance matrix adaptation evolution strategy (CMA-ES)
\cite{hansen2001completely} is a state-of-the-art evolutionary algorithm
for continuous blackbox functions (i.e. functions for which only the function
values are available for the search points\cite{hansen2016cma}). We used
the capability of CMA-ES's convergence to find a solution using small population
size, as compared to other evolutionary methods, for reducing the search
time.

For $N_{ops}$ number of operations in the search space and $N_{edges}$ edges
between nodes in the cell, $\alpha$ has $N_{ops}\times N_{edges}$ parameters.
These are modelled with a multivariate normal distribution
$\mathcal N(\textbf{m}, \textbf{C})$ with a mean vector, $\textbf{m}$, of
size $N_{ops}\times N_{edges}$, representing the parameters in $\alpha$,
and a covariance matrix, \textbf{C}, of size
$(N_{ops}\times N_{edges})\times(N_{ops}\times N_{edges})$. CMANAS is an
iterative process, where in each iteration/generation, a population of 
architectures (i.e. $\alpha$) with the population size $N_{pop}$ is sampled from
the normal distribution $\mathcal N(\textbf{m}, \textbf{C})$ according to the
following equation:
\begin{equation}
	\label{eq:sampling}
	\alpha_{_{1:N_{pop}}}^{(g+1)} \sim \textbf{m}^{(g)} +
	\sigma^{(g)}\mathcal N(0, \textbf{C}^{(g)})
\end{equation}
where $g=0,1,2...$ is the generation number and $\sigma$ is the step-size.
The fitness of the indivdual architecture in the population is then evaluated
using the trained OSM model (Section~\ref{subsect:performance}), and the
architectures are sorted in the decreasing order according to their fitness.
The top-$\mu$ individuals are used to update the mean, \textbf{m}, as follows:
\begin{equation}
	\label{eq:mean_update}
	\textbf{m}^{(g+1)} \gets \sum_{i=1}^{\mu} w_i\alpha_i^{(g+1)}
\end{equation}
where $\mu=N_{pop}/2$ and this process is also referred to as the
\textit{selection} and \textit{recombination} part of CMA-ES. The step-size,
$\sigma$, is then updated using a \textit{conjugate evolution path}, $p_{\sigma}$
which is calculated as the sum of the previous successive steps taken over a
number of generations. Its length is compared with the expected length under
random selection in order to decide whether to increase or decrease the $\sigma$,
as follows:
\begin{equation}
	\label{eq:sigma_update}
	\sigma^{(g+1)} \gets \sigma^{(g)} \times \exp\left(\frac{c_{\sigma}}{d_{\sigma}}
	\left(\dfrac{||p_{\sigma}^{(g)}||}{E||\mathcal N(0, \textbf{I})||}-1\right)
	\right)
\end{equation}
where $c_{\sigma}$ and $d_{\sigma}$ are the hyperparameters of CMA-ES. Lastly,
the covariance matrix, \textbf{C}, is updated using two terms: \textit{Rank}-$\mu$
update, ($Rank_{\mu}$) and \textit{Rank}-$1$ update, ($Rank_{1}$). $Rank_{\mu}$
is calculated using the previous generation covariance matrices with
higher weights given to recent generations. The $Rank_{1}$ is calculated using
another evolution path which is expressed as a sum of consecutive steps taken over
a number of generations. The whole process is summarized as follows:
\begin{equation}
	\label{eq:cov_update}
	\textbf{C}^{(g+1)} \gets (1+k)\textbf{C}^{(g)} + c_1 Rank_1^{(g)} +
	c_{\mu} Rank_{\mu}^{(g)}
\end{equation}
where $k, c_1$ and $c_{\mu}$ are the hyperparameters of CMA-ES. For a more detailed
analysis of the hyperparameters and the updates in CMA-ES, please refer to
\cite{hansen2016cma}.

\subsection{CMANAS}
As illustrated in Fig.~\ref{fig:arch_search}, CMANAS starts with
initializing the normal distribution $\mathcal N(\textbf{m}, \textbf{C})$.
\comm{The mean, $\textbf{m}$ is initialized with a zero vector and the
covariance matrix, $\textbf{C}$, is initialized with an identity matrix.}
This distribution is then used to sample a population of architectures
(using Eq. (\ref{eq:sampling})), which are then evaluated using the
trained OSM. In the evaluation process, we use an architecture-fitness table (AF
table) to save the fitness of the already evaluated architecture. For an individual
architecture in the population, we first check whether there is an entry
present in the AF table for this architecture. If it is already present, then the
entry in the AF table is returned. Otherwise, the fitness of the architecture is
evaluated using the trained OSM (Section~\ref{subsect:performance}), and the
AF table is updated with it. The normal distribution
$\mathcal N(\textit{m}, \textbf{C})$ and the step-size, $\sigma$, are updated
using CMA-ES according to the evaluated population. The updated distribution is
then re-sampled for the next generation population of architectures and repeat the
cycle for a certain number of generations, $N_{gen}$. The mean of the normal
distribution is returned as the searched architecture,
$\textbf{m}^{(N_{gen})}$ after $N_{gen}$ generations. The algorithm is summarized
in Algorithm~\ref{algo:CMANAS}.

\section{Experiments and Results}

\subsection{Baselines}
In order to illustrate the effectiveness of CMANAS, we compared the
architecture returned by CMANAS with the other architectures reported in various
peer-reviewed NAS methods. These peer-reviewed NAS methods are broadly classified
into five categories: architectures designed by human (reported as
\textit{manual}), RL based methods (reported as \textit{RL}), gradient-based
methods (reported as \textit{grad. based}), EA based methods (reported as
\textit{EA}) and \textit{others}. The \textit{others} include the random search
\cite{li2019random} and sequential model-based optimization
(SMBO) wherein the architecture is searched in the increasing order of complexity
of its structure. The effectiveness of the reported architectures are measured in
terms of the classification accuracy and the computational requirement,
measured in terms of search time on a single GPU {reported as GPU days}.

\subsection{Dataset Settings}
\label{subsect:data_settings}
Both \textbf{CIFAR-10} and \textbf{CIFAR-100} \cite{krizhevsky2009learning} have
50K training images and 10K testing images and are classified into 10 classes
and 100 classes respectively. \textbf{ImageNet} \cite{imagenet_cvpr09} is a
popular benchmark for image classification and contains 1.28 million training
images and 50K images test images, which are classified into 1K classes.
\textbf{ImageNet-16-120} \cite{chrabaszcz2017downsampled} is a down-sampled
version of ImageNet wherein the images in the original ImageNet dataset are
downsampled to $16\times16$ pixels with 120 classes to construct the
ImageNet-16-120 dataset. The settings used for the datasets in \textbf{S1} are as
follows:
\begin{itemize}
	\item \textit{CIFAR-10}: We followed \cite{liu2018darts2} and split 50K
	training images into two sets of size 25K each, with one set acting as the
	training set and the other set as the validation set.
	\item \textit{CIFAR-100}: We followed \cite{lu2020multi} and split 50K
	training images into two sets. One set of size 40K images becomes the
	training set and the other set of size 10K images becomes the validation set.
\end{itemize}
We followed the settings used in \cite{Dong2020NAS-Bench-201} for the datasets in
\textbf{S2} which are as follows:
\begin{itemize}
	\item \textit{CIFAR-10}: The same settings as those used for S1 is used here
	as well.
	\item \textit{CIFAR-100}: The 50K training images remains as the training set
	and the 10K testing images are split into two sets of size 5K each, with one
	set acting as the validation set and the other set as the test set.
	\item \textit{ImageNet-16-120}: It has 151.7K training images, 3K validation
	images and 3K test images.
\end{itemize}
The training set is used for training the OSM (Fig.~\ref{fig:training})
and the validation set is used for estimating the fitness of the sampled
architecture during the search process (Section~\ref{subsect:performance}).

\subsection{Implementation Details}
\subsubsection{\textbf{Search Space 1 (S1)}}
This is similar to that used in DARTS \cite{liu2018darts2}, which allows us to
compare the performance of CMANAS with other NAS methods. Here, we search
for both normal and reduction cells in Fig.~\ref{fig:macro_arch}(a) wherein each
cell has seven nodes with first two nodes being the output from previous cells and
last node as output node, resulting in 14 edges among them. There are eight
operations considered in S1 which are as follows: $3\times3$ and $5\times5$
dilated separable convolutions, $3\times3$ and $5\times5$ separable convolutions,
$3\times3$ max pooling, $3\times3$ average pooling, skip connect and zero.
Therefore, an architecture is represented by two $14\times8$ matrices, one each
for a normal cell and a reduction cell. The values in these two matrices are
modelled with a multivariate normal distribution $\mathcal N(\textbf{m},
\textbf{C})$ with a mean vector, $\textbf{m}$, of size $224$, and a covariance
matrix, \textbf{C}, of size $224\times224$.

\subsubsection{\textbf{Search Space 2 (S2)}}
\label{subsubsect:training_settings}
\begin{figure}[t]
	\centering
	\begin{center}
		\includegraphics[width=\linewidth]{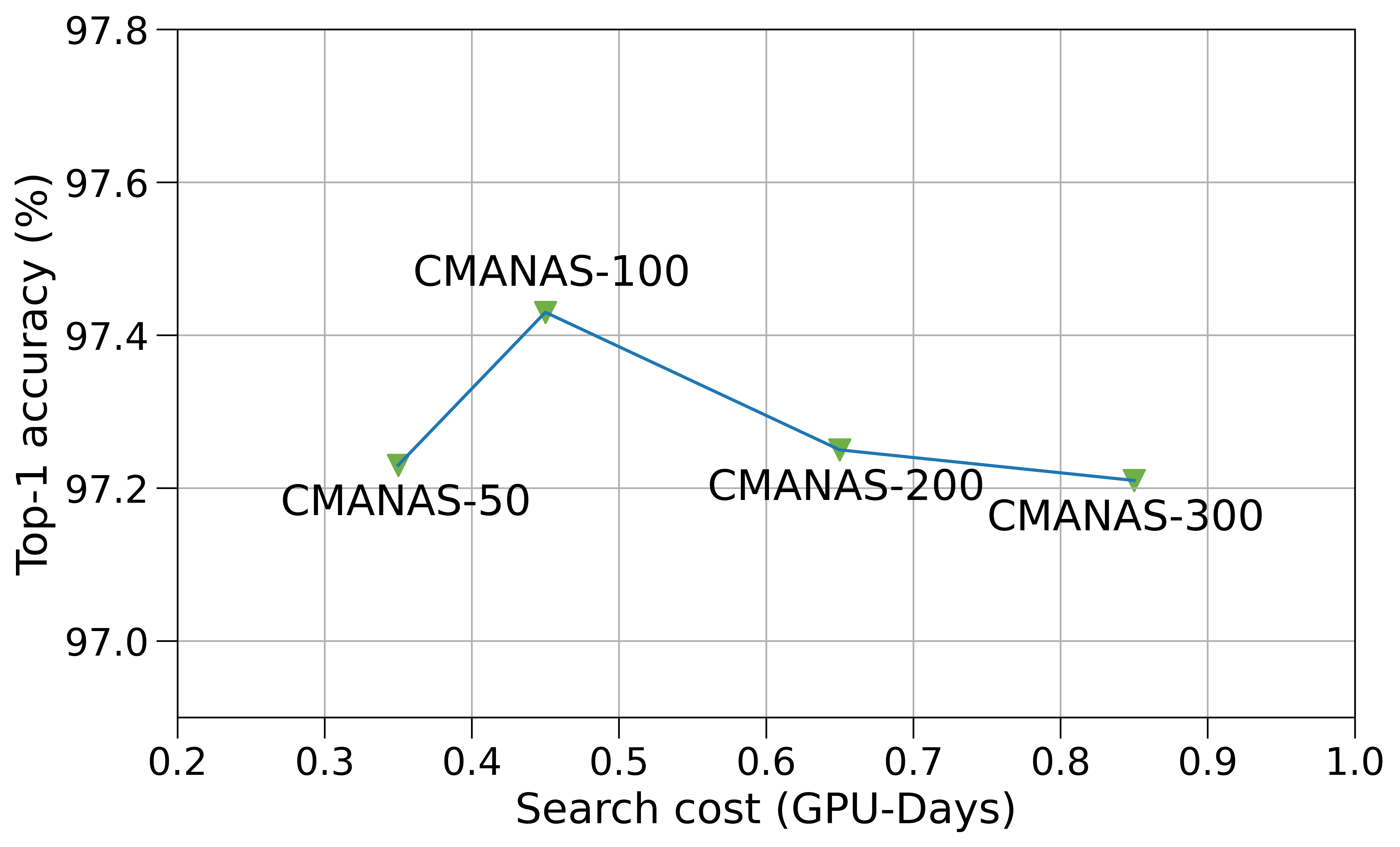}
	\end{center}
	\caption{Top1 accuracy of architecture searches performed with OSM trained for
		50, 100, 200 and 300 epochs in S1 on CIFAR-10 dataset. The number after
		CMANAS specifies the number of epochs.}
	\label{fig:training_epochs}
\end{figure}
This is a smaller search space with a total of 15,625 architectures in the search
space and is similar to that used in NAS-Bench-201 \cite{Dong2020NAS-Bench-201},
where we only search for the normal cell in Figure~\ref{fig:macro_arch}(a).
NAS-Bench-201 provides a unified benchmark for almost any up-to-date NAS
algorithms by providing the results of each architecture in the search space on
CIFAR-10, CIFAR-100 and ImageNet16-120. It provides an API that can be used to
query accuracies on both validation and test sets for all the architectures in the
search space. The API provides two types of accuracies for each architecture, i.e.
accuracy after training the architecture for 12 epochs and 200 epochs. The
accuracies of the architectures after 200 epochs are used as the performance
measurement of various NAS algorithms. NAS-Bench-201 \cite{Dong2020NAS-Bench-201}
(i.e. S2) provides the search results for two types of NAS methods: 
\textit{weight sharing} based and \textit{non-weight sharing} based. In the weight
sharing based NAS methods, all the architectures in the search space share their
weights to reduce the search time (e.g. 
\cite{pmlr-v80-pham18a}\cite{liu2018darts2}\cite{dong2019searching}\cite{dong2019one}\cite{li2019random}).
In the non-weight sharing based NAS methods (e.g.
\cite{real2019regularized}\cite{bergstra2012random}\cite{williams1992simple}
\cite{falkner2018bohb}), the architectures in the search space do not share their
weights, and during the architecture search, the performance of each architecture
is evaluated on the basis of the accuracy on the validation data after training
for 12 epochs which is provided by the API.

Here, each cell has four nodes with the first node as the input node and last node
as the output node, resulting in six edges among them. The five operations
considered in S2 are as follows: $1\times1$ and $3\times3$ convolutions,
$3\times3$ average pooling, skip connect and zero. Therefore, an architecture is
represented by a $6\times5$ matrix for the normal cell. The values in the matrix
are modelled with a multivariate normal distribution,
$\mathcal N(\textbf{m}, \textbf{C})$, with a mean vector, $\textbf{m}$, of size
$30$, and a covariance matrix, \textbf{C}, of size $30\times30$.

\subsubsection{\textbf{Training Settings}}
\begin{figure*}[t]
	\centering
	\subfloat[]{
		\label{subfig:val_acc_mean_s1}
		\includegraphics[width=0.49\linewidth]{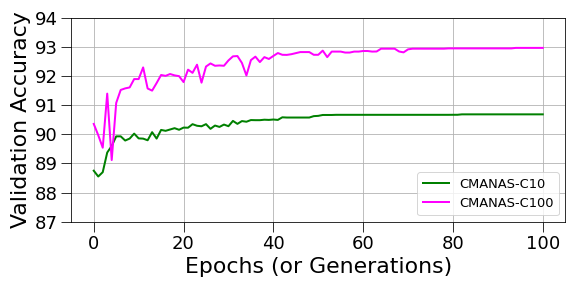}
	}
	\subfloat[]{
		\label{subfig:val_acc_mean_s2}
		\includegraphics[width=0.49\linewidth]{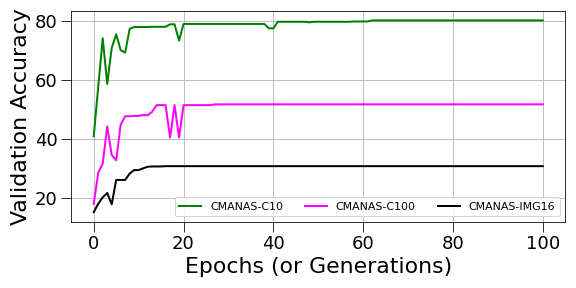}
	}
	\caption{Plot of validation accuracy/fitness of the mean, $\textbf{m}$, of the
		normal distribution, $\mathcal N(\textbf{m}, \textbf{C})$ at each
		epoch/generation in		
		(a) S1 on CIFAR-10 (denoted as CMANAS-C10) and CIFAR-100 (denoted as
		CMANAS-C100).
		(b) S2 on CIFAR-10 (denoted as CMANAS-C10), CIFAR-100 (denoted as
		CMANAS-C100) and ImageNet16-120	(denoted as CMANAS-IMG16).
		All the plots are averaged over all three runs in both S1 and S2.
	}
	\label{fig:val_acc_mean}
\end{figure*}
\begin{figure*}[h]
	\centering
	\subfloat[]{
		\label{subfig:searched_cells_c10}
		\includegraphics[width=0.45\linewidth]{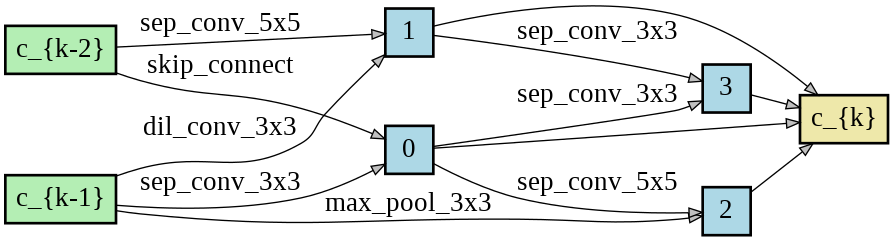}
		\qquad
		\includegraphics[width=0.45\linewidth]{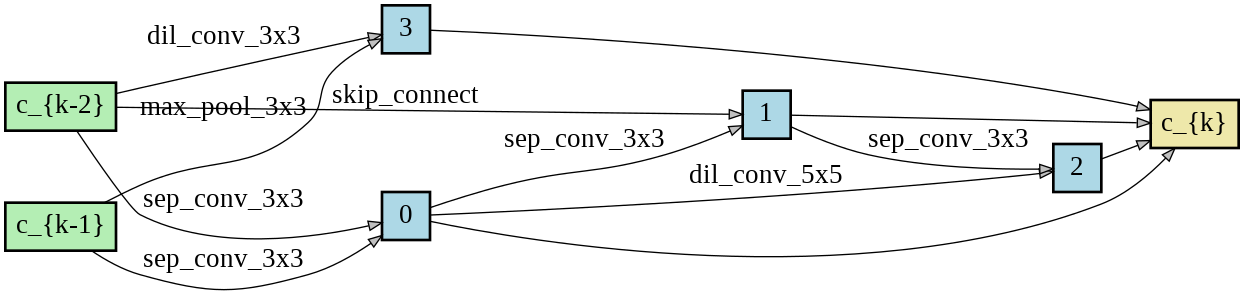}
	}
	\qquad
	\subfloat[]{
		\label{subfig:searched_cells_c100}
		\includegraphics[width=0.45\linewidth]{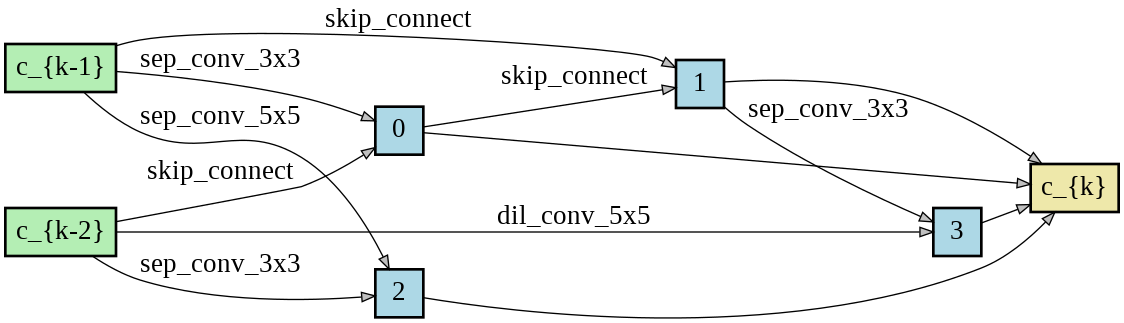}
		\qquad
		\includegraphics[width=0.45\linewidth]{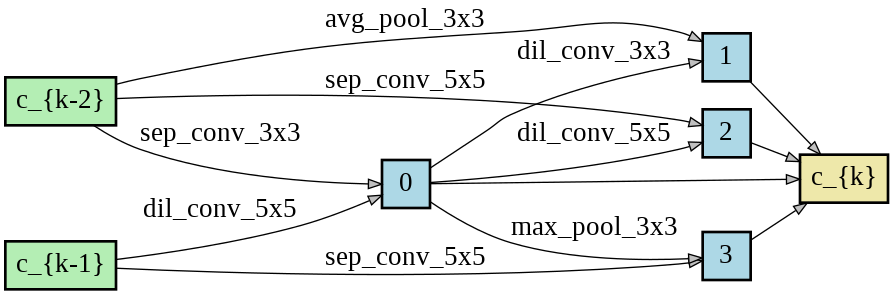}
	}
	\caption{Normal cell on the left side and reduction cell in the right side.
		(a) Cells discovered by CMANAS-C10A
		(b) Cells discovered by CMANAS-C100A}
	\label{fig:searched_cells}
\end{figure*}
The training process is executed two times in our method which are as follows:
\begin{itemize}
	\item \textit{One shot model (OSM) training}: In general, the OSM suffers from
	high memory requirements which makes it difficult to fit it in a single GPU.
	For S1, we follow \cite{liu2018darts2}\cite{li2019random} and use a
	smaller OSM, called \textit{proxy model} which is created with 8 stacked
	cells and 16 initial channels. It is then trained with SGD for 100 epochs on
	both CIAFR-10 and CIFAR-100 with the same settings i.e. batch size of 96,
	weight 	decay $\lambda=3\times10^{-4}$, cutout\cite{devries2017improved},
	initial learning rate $\eta_{max}=0.025$ (annealed down to 0 by using a cosine
	schedule without restart\cite{DBLP:conf/iclr/LoshchilovH17}) and momentum
	$\rho=0.9$. For	\textit{S2}, we do not use a proxy model as the size of the
	OSM is sufficiently small to be fitted in a single GPU. For training, we
	follow the same settings as those used in S1 for CIFAR-10, CIFAR-100 and
	ImageNet16-120 except batch size of 256.
	
	\item \textit{Architecture evaluation}: Here, the discovered architecture,
	$\mathcal A$ (i.e. discovered cells), at the end of the architecture search is
	trained on the dataset to evaluate its performance for comparing with other
	NAS methods. For \textit{S1}, we follow the training settings used in DARTS
	\cite{liu2018darts2}. Here, $\mathcal A$ is created with 20 stacked cells and
	36 initial channels for both CIFAR-10 and CIFAR-100 datasets. It is then
	trained for 600 epochs on both the datasets with the same settings as the ones
	used in the OSM training above. Following recent works
	\cite{pmlr-v80-pham18a}\cite{real2019regularized}\cite{zoph2018learning}
	\cite{liu2018darts2}\cite{liu2018progressive}, we use an auxiliary tower with
	0.4 as its	weights, path dropout probability of 0.2 and
	cutout\cite{devries2017improved} for additional enhancements. For
	ImageNet, $\mathcal A$ is created with 14 cells and 48 initial channels in the
	mobile setting, wherein the input image size is 224 x 224 and the number of
	multiply-add operations in the model is restricted to less than 600M. It is
	trained on 8 NVIDIA V100 GPUs by following the training settings used in
	\cite{chen2019progressive}.
\end{itemize}
All the above trainings were performed on a single Nvidia RTX 3090 GPU except the
one on ImageNet. The number of epochs for training the ONS was chosen to be 100
as we found that the performance of the architecture search increases from 50
epochs to 100 epochs and then deteriorates upon further increase in the number of
epochs because of the overfitting of the OSM (as shown in
Fig.~\ref{fig:training_epochs}).

\subsubsection{\textbf{Architecture Search Settings}}
The multivariate normal distribution,
$\mathcal N(\textbf{m}, \textbf{C})$, that is used to model the architecture
parameter, $\alpha$, is initialized with its mean, $\textbf{m}$ equal to the zero
vector. This results in the assignment of equal weights to all the operations in
all the edges because of the normalization by softmax (Eq.~\ref{eq:ons}). As
recommended in \cite{hansen2016cma}, we initialize the covariance matrix,
$\textbf{C}$, with an identity matrix and the population size, $N_{pop}$ to
$4+\left \lfloor{3\times\ln(n)}\right \rfloor$, where $n$ is the size of
the mean, $\textbf{m}$. Therefore, for S1, $N_{pop}=20$ and for S2,
$N_{pop}=14$. The other hyperparameters of CMA-ES are initialized to
their default values as per the recommendation in \cite{hansen2016cma}.
We run CMANAS for 100 epochs/generations as we observed that the algorithm
converged well before 100 epochs for both S1 and S2 (as shown in
Fig.~\ref{fig:val_acc_mean}). All the architecture searches were performed on
a single Nvidia RTX 3090 GPU. All the codes were implemented using the deep
learning framework, PyTorch\cite{NEURIPS2019_9015}.

\begin{table*}[t]
	\caption{Comparison of CMANAS with other NAS methods in S1 in terms
		of test accuracy (higher is better) on (a) CIFAR-10 (b) CIFAR-100 
		and (c) ImageNet.}
	\begin{minipage}{0.5\textwidth}
		\centering
		\label{table:c10_s1}
		\begin{tabular}{l|c|c|c|c}
			
			\hline
			\bf{Architecture}& \multicolumn{1}{c|}{\bf{Top-1}} & \bf{Params} & \bf{GPU} &\bf{Search} \\
			&\textbf{Acc.} (\%)&(M)& \bf{Days} &\bf{Method}\\
			\hline
			ResNet  \cite{he2016deep}  & 95.39 & 1.7 & - & manual\\
			DenseNet-BC  \cite{huang2017densely}  & 96.54 & 25.6 & - & manual\\
			\hline
			
			PNAS \cite{liu2018progressive}  & 96.59 & 3.2 & 225 & SMBO\\
			RSPS\cite{li2019random}         & 97.14 & 4.3 &2.7& random\\
			\hline
			
			NASNet-A   \cite{zoph2018learning}     & 97.35 & 3.3  &1800& RL\\
			ENAS       \cite{pmlr-v80-pham18a}    & 97.14  & 4.6  &0.45& RL\\
			\hline
			DARTS      \cite{liu2018darts2}     & 97.24 & 3.3 &4& grad. based\\
			GDAS       \cite{dong2019searching} & 97.07 & 3.4 &0.83& grad. based\\
			SNAS       \cite{xie2018snas}       & 97.15 & 2.8&1.5& grad. based\\
			SETN       \cite{dong2019one}       & 97.31 & 4.6 &1.8& grad. based\\
			\hline
			
			AmoebaNet-A \cite{real2019regularized}  & 96.66 & 3.2 &3150& EA\\
			Large-scale Evo.\comm{$^\dagger$} \cite{real2017large}  & 94.60 & 5.4 &2750& EA\\
			Hierarchical Evo.\comm{$^\dagger$} \cite{liu2018hierarchical}  & 96.25 & 15.7 &300& EA\\
			CNN-GA \cite{sun2020automatically}  & 96.78 & 2.9 &35& EA\\
			CGP-CNN \cite{suganuma2017genetic}  & 94.02 & 1.7 &27& EA\\
			AE-CNN \cite{sun2019completely}  & 95.7 & 2.0 &27& EA\\
			NSGANetV1-A2 \cite{lu2020multi}  & 97.35 & 0.9 &27& EA\\
			AE-CNN+E2EPP \cite{sun2019surrogate}  & 94.70 & 4.3 &7& EA\\
			NSGA-NET \cite{lu2019nsga}  & 97.25 & 3.3 &4& EA\\
			\comm{\hline
				EvNAS-A (Ours)                       & 97.53 & 3.6 & 3.83 & EA\\
				EvNAS-B (Ours)                       & 97.38 & 3.8 & 3.83 & EA\\
				EvNAS-C (Ours)                       & 97.37 & 3.4 & 3.83 & EA\\}
			\hline
			\bf{CMANAS-C10A}     & \bf{97.44} & \bf{3.8} & \bf{0.45} & \bf{EA}\\
			CMANAS-C10B              & 97.35      & 3.2      & 0.45      & EA\\
			CMANAS-C10C              & 97.35      & 3.3      & 0.45      & EA\\
			\hline
			CMANAS-C10rand              & 97.11      & 3.11 & 0.66 & random\\
			\hline
		\end{tabular}
		\vskip2ex \normalsize (a) CIFAR-10
		\vspace*{0.2cm}
	\end{minipage}
	\begin{minipage}{0.5\textwidth}
		\label{table:c100_s1}
		\centering
		\begin{tabular}{l|c|c|c|c}
			\hline
			\bf{Architecture}& \multicolumn{1}{c|}{\bf{Top-1}} & \bf{Params} & \bf{GPU} &\bf{Search} \\
			&\textbf{Acc.} (\%)& (M) & \bf{Days} &\bf{Method}\\
			\hline
			ResNet  \cite{he2016deep}  & 77.90 & 1.7 & - & manual\\
			DenseNet-BC         \cite{huang2017densely}  & 82.82   & 25.6 & - & manual\\
			\hline
			PNAS            \cite{liu2018progressive}  & 80.47 & 3.2 & 225 & SMBO\\
			\hline
			
			\comm{NASNet-A   \cite{zoph2018learning}     & -     & 3.3  &1800& RL\\}
			MetaQNN \cite{baker2017designing}      & 72.86      & 11.2  &90& RL\\
			ENAS       \cite{pmlr-v80-pham18a}     & 80.57      & 4.6  &0.45& RL\\
			\hline
			
			DARTS      \cite{liu2018darts2}     & 82.46 & 3.3 &4& grad. based\\
			GDAS       \cite{dong2019searching} & 81.62 & 3.4 &0.83& grad. based\\
			SETN       \cite{dong2019one}       & 82.75 & 4.6 &1.8& grad. based\\
			\hline
			
			AmoebaNet-A \cite{real2019regularized}  & 81.07 & 3.2 &3150& EA\\
			Large-scale Evo.\comm{$^\dagger$} \cite{real2017large}  & 77.00 & 40.4 &2750& EA\\
			CNN-GA \cite{sun2020automatically}  & 79.47 & 4.1 &40& EA\\
			AE-CNN \cite{sun2019completely}  & 79.15 & 5.4 &36& EA\\
			NSGANetV1-A2 \cite{lu2020multi} & 82.58 & 0.9 &27& EA\\
			Genetic CNN \cite{xie2017genetic}  & 70.95 & - &17& EA\\
			AE-CNN+E2EPP \cite{sun2019surrogate}  & 77.98 & 20.9 &10& EA\\
			NSGA-NET \cite{lu2019nsga}  & 79.26 & 3.3 &8& EA\\
			\comm{EENA       \cite{zhu2019eena}  & 82.29 & 8.47 & 0.65 & EA\\
			\hline
				EvNAS-A (Ours)                 & 83.63& 3.6 & 3.83 & EA\\
				EvNAS-B (Ours)                 & 83.49 & 3.8 & 3.83 & EA\\
				EvNAS-C (Ours)                 & 83.14 & 3.4 & 3.83 & EA\\
			}
			\hline
			\bf{CMANAS-C100A}        & \bf{83.24} & \bf{3.4} & \bf{0.60} & \bf{EA}\\
			CMANAS-C100B            & 83.09   & 3.47      & 0.63 & EA\\
			CMANAS-C100C            & 82.73   & 2.97      & 0.62 & EA\\
			\hline
			CMANAS-C100rand         & 82.35      & 3.17 & 0.67 & random\\
			\hline
		\end{tabular}
		\vskip2ex \normalsize (b) CIFAR-100
	\end{minipage}

	\begin{minipage}{\textwidth}
		\centering
		\label{table:imagenet_s1}
		\begin{tabular}{l|c|c|c|c|c|c}
			\hline
			\bf{Architecture} & \multicolumn{2}{c|}{\bf{Test Accuracy (\%)}} & \bf{Params} &+$\times$& \bf{Search Time} &\bf{Search} \\
			& \bf{top 1} & \bf{top 5} & (M) & (M) & (GPU Days) & \bf{Method} \\
			\hline
			MobileNet-V2  \cite{sandler2018mobilenetv2}&72.0& 91.0 & 3.4 & 300 & - & manual\\ 
			\hline
			PNAS \cite{liu2018progressive}  &74.2& 91.9    & 5.1 & 588 & 225 & SMBO\\
			\hline
			
			NASNet-A             \cite{zoph2018learning}    & 74.0 & 91.6      & 5.3 & 564 &1800& RL\\
			NASNet-B             \cite{zoph2018learning}    & 72.8  & 91.3    & 5.3 & 488 &1800& RL\\
			NASNet-C             \cite{zoph2018learning}    & 72.5  & 91.0     & 4.9 & 558 &1800& RL\\
			\hline
			
			DARTS  \cite{liu2018darts2}           & 73.3  & 91.3 & 4.7 & 574 &4& grad. based\\
			GDAS   \cite{dong2019searching}       & 74.0  & 91.5 & 5.3 & 581 &0.83& grad. based\\
			SNAS   \cite{xie2018snas}             & 72.7  & 90.8 & 4.3 & 522 &1.5& grad. based\\
			SETN   \cite{dong2019one}             & 74.3  & 92.0 & 5.4 & 599 &1.8& grad. based\\
			\hline
			
			AmoebaNet-A          \cite{real2019regularized} & 74.5 & 92.0 & 5.1 & 555 &3150& EA\\
			AmoebaNet-B          \cite{real2019regularized} & 74.0 & 91.5 & 5.3 & 555 &3150& EA\\
			AmoebaNet-C          \cite{real2019regularized} & 75.7 & 92.4 & 6.4 & 570 &3150& EA\\
			NSGANetV1-A2         \cite{lu2020multi} & 74.5 & 92.0 & 4.1 & 466 &27& EA\\
			\comm{FairNAS-A          \cite{chu2019fairnas} & 75.3 & 92.4 & 4.6 & 388 &12& EA\\
				FairNAS-B          \cite{chu2019fairnas} & 75.1 & 92.3 & 4.5 & 345 &12& EA\\
				FairNAS-C          \cite{chu2019fairnas} & 74.7 & 92.1 & 4.4 & 321 &12& EA\\
				\hline
				
				EvNAS-A (Ours)                                     & 75.6 & 92.6 & 5.1 & 570 & 3.83 & EA\\
				EvNAS-B (Ours)                                     & 75.6 & 92.6 & 5.3 & 599 & 3.83 & EA\\
				EvNAS-C (Ours)                                     & 74.9 & 92.2 & 4.9 & 547 & 3.83 & EA\\
			}
			\hline
			\bf{CMANAS-C10A} & \bf{75.3} & \bf{92.6} & \bf{5.3} & \bf{589} & \bf{0.45} & \bf{EA}\\
			CMANAS-C100A                & 74.8 & 92.1 & 4.8 & 531 & 0.60 & EA\\
			\hline
		\end{tabular}
		\vskip2ex \normalsize (c) ImageNet
	\end{minipage}
\end{table*}

\subsection{Results}
\label{subsect:results}
\subsubsection{\textbf{Search Space 1 (S1)}}
\label{subsect:results_S1}
\begin{figure}[t]
	\centering
	\begin{center}
		\includegraphics[width=\linewidth]{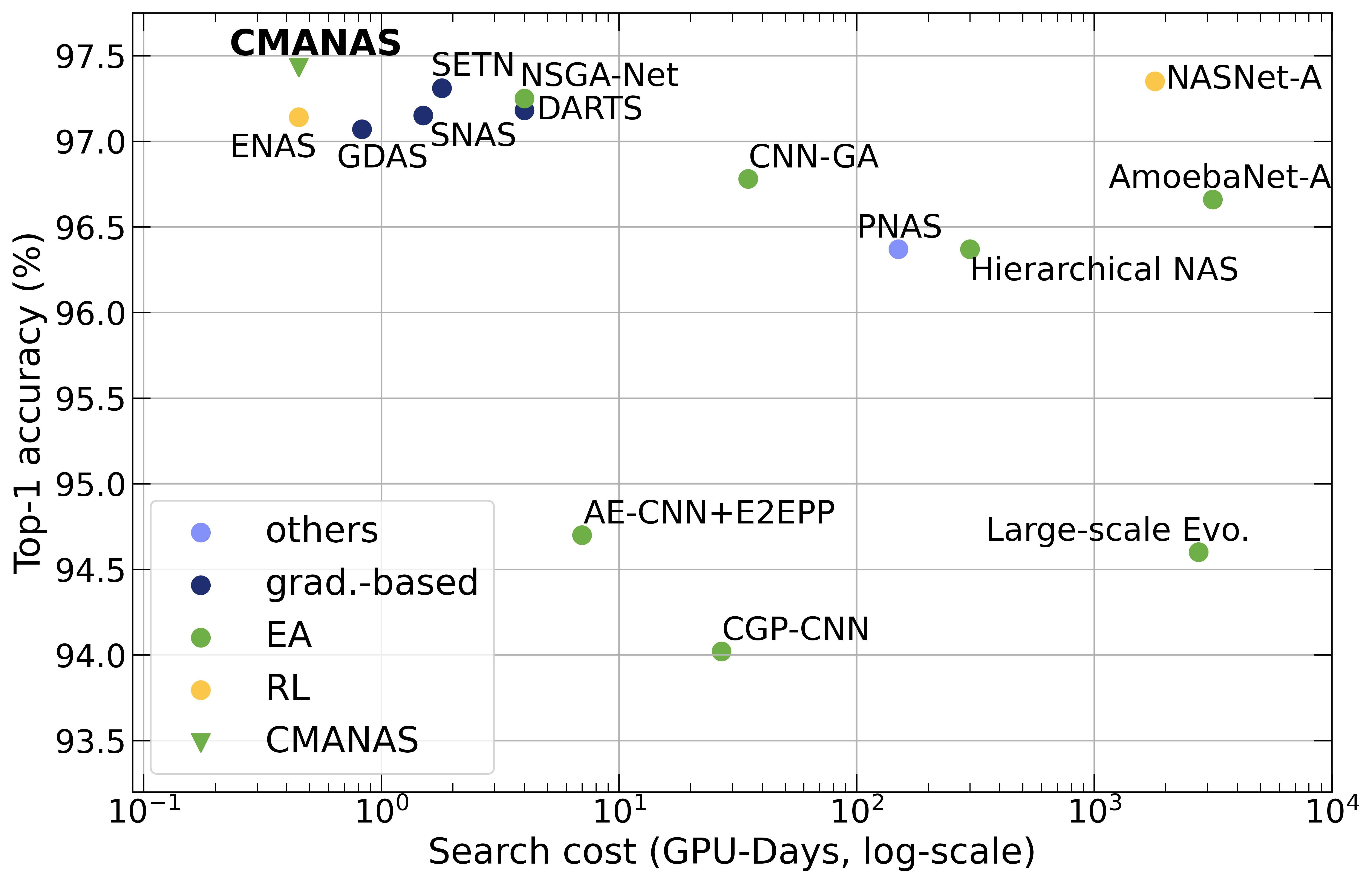}
	\end{center}
	\caption{Search cost comparision of CMANAS with the other NAS algorithms
		in S1.}
	\label{fig:search_cost}
\end{figure}

We performed three architecture searches on CIFAR-10 with different random number
seeds; their results are provided in Table~\ref{table:c10_s1}(a) as CMANAS-C10A,
CMANAS-C10B and CMANAS-C10C. We also performed another three architecture search
on CIFAR-100 with different random number seeds; their results are provided in
Table~\ref{table:c100_s1}(b) as CMANAS-C100A, CMANAS-C100B and CMANAS-C100C. The
results show that the cells discovered by CMANAS on CIFAR-10 and CIFAR-100
achieve better results than those by human designed, RL based, gradient based and
EA based methods while using significantly less computational time. We compared
the computation time (or \textit{search cost}) spent on the \textit{GPU days}, of
CMANAS with that for the other NAS methods (as shown in 
Fig.~\ref{fig:search_cost}). GPU days for any NAS method is calculated by
multiplying the number of GPUs used in the NAS method by the execution time
(reported in units of days). A single run of CMANAS on CIFAR-10 and CIFAR-100
took 0.45 and 0.6 GPU days (including the training time of the OSM and the
architecture search time using the trained OSM on the dataset), respectively. For
comparison with other NAS methods, the search cost on CIFAR-10 for CMANAS was
used, as it is the most common search cost used in most of the NAS methods. From
Fig.~\ref{fig:search_cost}, we observe that CMANAS is able to achieve better
results than the previous evolution based methods
AmoebaNet\cite{real2019regularized}, Large-scale Evolution\cite{real2017large}
and Hierarchical Evolution\cite{liu2018hierarchical}
while using \textbf{10x} to \textbf{1000x} less GPU days. The top cells
discovered by CMANAS on CIFAR-10 and CIFAR-100 (i.e. CMANAS-C10A, CMANAS-C100A)
are shown in Fig.~\ref{fig:searched_cells}. The cells discovered by the other runs
of CMANAS on CIFAR-10 and CIFAR-100 are provided in the supplementary.

We followed
\cite{pmlr-v80-pham18a}\cite{real2019regularized}\cite{zoph2018learning}\cite{liu2018darts2}\cite{liu2018progressive} to compare the transfer capability of CMANAS
with that of the other NAS methods, wherein the discovered architecture on a
dataset was transferred to another dataset (i.e. ImageNet) by retraining the
architecture from scratch on the new dataset. The best discovered architectures
from the architecture search on CIFAR-10 and CIFAR-100 (i.e. CMANAS-C10A and
CMANAS-C100A) were then evaluated on the ImageNet dataset in mobile setting and the
results are provided in Table~\ref{table:imagenet_s1}(c). The results show that the
cells discovered by CMANAS on CIFAR-10 and CIFAR-100 can be successfully
transferred to ImageNet, achieving better results than those of human
designed, RL based, gradient based and EA based methods while using significantly
less computational time. \comm{Notably, CMANAS is able to achieve better
result than previous state-of-the-art evolution based methods
\cite{real2019regularized} while using significantly less computational time.}

\begin{table*}[t]
	\caption{Comparison of CMANAS with other NAS methods on NAS-Bench-201 (i.e. S2)
		\cite{Dong2020NAS-Bench-201} with mean $\pm$ std. accuracies on CIFAR-10,
		CIFAR-100 and ImageNet16-120 (higher is better). The first block
		compares CMANAS with other weight sharing based NAS methods. The second
		block compares CMANAS with other non-weight sharing based NAS methods.
		Optimal in the third block refers to the best architecture accuracy for
		each dataset. Search times are given for a CIFAR-10 search on a single
		GPU.}
	\label{table:NAS201}
	\centering
	\begin{tabular}{l|c|cc|cc|cc|c}
		\hline
		\bf{Method}& \bf{Search} & \multicolumn{2}{c|}{\bf{CIFAR-10}} & \multicolumn{2}{c|}{\bf{CIFAR-100}} & \multicolumn{2}{c|}{\bf{ImageNet-16-120}} & \bf{Search}\\
		&(seconds)& \it{validation} & \it{test} & \it{validation} & \it{test} & \it{validation} & \it{test}
		& \bf{Method}\\
		\hline
		RSPS\comm{$^\dagger$} \cite{li2019random} & $7587.12$ &$84.16\pm1.69$&$87.66\pm1.69$&$59.00\pm4.60$&$58.33\pm4.64$& $31.56\pm3.28$&$31.14\pm3.88$& random\\
		
		DARTS-V1 \cite{liu2018darts2} &$10889.87$ & $39.77\pm0.00$ & $54.30\pm0.00$ & $15.03\pm0.00$ &$15.61\pm0.00$ & $16.43\pm0.00$&$16.32\pm0.00$& grad. based\\
		
		DARTS-V2 \cite{liu2018darts2} & $29901.67$ & $39.77\pm0.00$ & $54.30\pm0.00$ & $15.03\pm0.00$ &$15.61\pm0.00$& $16.43\pm0.00$&$16.32\pm0.00$& grad. based\\
		
		GDAS\comm{$^\dagger$} \cite{dong2019searching} & $28925.91$ & $90.00\pm0.21$ & $93.51\pm0.13$ & $71.14\pm0.27$ &$70.61\pm0.26$& $41.70\pm1.26$&$41.84\pm0.90$& grad. based\\
		
		SETN\comm{$^\dagger$}\cite{dong2019one} & $31009.81$ &$82.25\pm5.17$&$86.19\pm4.63$&$56.86\pm7.59$&$56.87\pm7.77$ &$32.54\pm3.63$&$31.90\pm4.07$& grad. based\\
		
		ENAS\comm{$^\dagger$} \cite{pmlr-v80-pham18a} & $13314.51$ &$39.77\pm$0.00&$54.30\pm0.00$&$15.03\pm00$&$15.61\pm0.00$& $16.43\pm0.00$&$16.32\pm0.00$& RL\\
		
		
		\bf{CMANAS} &\bf{13896}&\bf{89.06}$\pm$\bf{0.4}&\bf{92.05}$\pm$\bf{0.26}&\bf{67.43}$\pm$\bf{0.42}&\bf{67.81}$\pm$\bf{0.15}& \bf{39.54}$\pm$\bf{0.91}&\bf{39.77}$\pm$\bf{0.57}&\bf{EA}\\
		
		\hline
		AmoebaNet\comm{REA} \cite{real2019regularized} &$0.02$ & $91.19\pm0.31$ & $93.92\pm0.30$ & $71.81\pm1.12$ &$71.84\pm0.99$ & $45.15\pm0.89$&$45.54\pm1.03$& EA\\
		
		RS \cite{bergstra2012random} &$0.01$ & $90.93\pm0.36$ & $93.70\pm0.36$ & $70.93\pm1.09$ &$71.04\pm1.07$ & $44.45\pm1.10$&$44.57\pm1.25$& random\\
		
		REINFORCE \cite{williams1992simple} &$0.12$ & $91.09\pm0.37$ & $93.85\pm0.37$ & $71.61\pm1.12$ &$71.71\pm1.09$ & $45.05\pm1.02$&$45.24\pm1.18$& RL\\
		
		BOHB \cite{falkner2018bohb} &$3.59$ & $90.82\pm0.53$ & $93.61\pm0.52$ & $70.74\pm1.29$ &$70.85\pm1.28$ & $44.26\pm1.36$&$44.42\pm1.49$& grad. based\\
		
		\bf{CMANAS-h12-Ep25} &\bf{3.64}&\bf{91.23}$\pm$\bf{0.40}&\bf{94.00}$\pm$\bf{0.39}&\bf{72.16}$\pm$\bf{1.19}&\bf{72.11}$\pm$\bf{1.10}& \bf{45.69}$\pm$\bf{0.84}&\bf{45.7}$\pm$\bf{0.79}&\bf{EA}\\
		
		CMANAS-h12-Ep100 & 7.12 &91.28$\pm$0.35 &94.06$\pm$0.34 &72.40$\pm$0.96 &72.26$\pm$0.84 & 45.74$\pm$0.86 & 45.69$\pm$0.88 & EA\\
		
		\hline
		ResNet & N/A &$90.83$&$93.97$&$70.42$&$70.86$&$44.53$&$43.63$& manual\\
		Optimal & N/A &$91.61$&$94.37$&$73.49$&$73.51$&$46.77$&$47.31$& N/A\\
		\hline
		
	\end{tabular}
\end{table*}

\subsubsection{\textbf{Search Space 2 (S2)}}
\label{subsect:results_S2}
\begin{figure}[h]
	\centering
	\begin{center}
		\includegraphics[width=\linewidth]{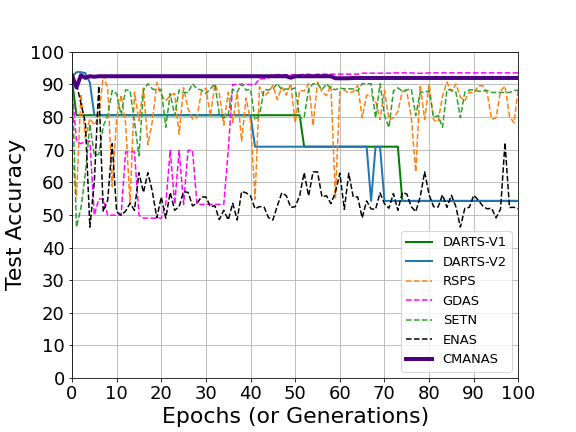}
	\end{center}
	\caption{Comparision of the weight sharing based CMANAS with other weight
		sharing based NAS methods in terms of the test accuracy of the derived
		architecture evaluated on CIFAR-10 at each epoch for the search space S2.}
	\label{fig:cmanas_NAS201_comparision}
\end{figure}

We performed architecture search on the CIFAR-10, CIFAR-100 and ImageNet-16-120
datasets for both the types of NAS methods given in S2\cite{Dong2020NAS-Bench-201}:
\begin{itemize}
	\item \textit{Weight sharing} based NAS: Here, the architecture evaluation
	in CMANAS used the trained OSM (as discussed in
	Section~\ref{subsect:performance}). Following \cite{Dong2020NAS-Bench-201},
	we perform the architecture search three times on all the three datasets and
	compared the results with those of other weight sharing based NAS mathods
	because of the weight sharing nature of the OSM. The results are reported as
	CMANAS in Table~\ref{table:NAS201}.
	
	\item \textit{Non-Weight sharing} based NAS: Here, the fitness of the 
	architecture was evaluated to be the accuracy on the validation data after
	training for 12 epochs, which is provided by the API in S2. Following
	\cite{Dong2020NAS-Bench-201}, we performed the architecture search 500 times on
	all the three datasets for 25 epochs each and compared the results with those
	of the other non-weight sharing based NAS methods; the corresponding results
	are reported as CMANAS-h12-Ep25 in Table~\ref{table:NAS201}. We also performed
	another architecture search 500 times on all the three datasets for 100 epochs
	each and reported the corresponding results as CMANAS-h12-Ep100 in
	Table~\ref{table:NAS201}; we found no significant improvement over the 25
	epoch version.
\end{itemize}
The results show that CMANAS outperforms most of the weight sharing NAS methods
except GDAS \cite{dong2019searching}. However, GDAS performs worse when the
size of the search space increases as can be seen for S1 in
Table~\ref{table:c10_s1}. CMANAS also dominates the non-weight sharing based NAS
methods. Notably, the non-weight sharing based CMANAS (i.e. CMANAS-h12-Ep25) also
dominates the weight sharing based CMANAS, which shows that the trained OSM
provides a noisy estimate of the fitness/performance of an architecture.

\textbf{CMANAS vs Gradient-Based methods:} In
Fig.~\ref{fig:cmanas_NAS201_comparision}, we compare the progression of the
search of the weight sharing based CMANAS with that of the other weight sharing
based NAS methods. The gradient-based method like DARTS\cite{liu2018darts2}
suffers from overfitting problem wherein it converges to operations that give
faster gradient descent, i.e. \textit{skip-connect} operation due to its
parameter-less nature as reported in \cite{chen2019progressive}
\cite{Zela2020Understanding}\cite{Dong2020NAS-Bench-201}. This leads to higher
number of \textit{skip-connect} in the final discovered cell, a local optimum (as
shown in Fig.~\ref{fig:cmanas_NAS201_comparision}). In contrast, CMANAS does not
get stuck to a local optimum architecture due to its stochastic nature.

\begin{figure*}[t]
	\centering
	\subfloat[]{
		\label{subfig:arch_search_viz_a}
		\includegraphics[width=\linewidth]{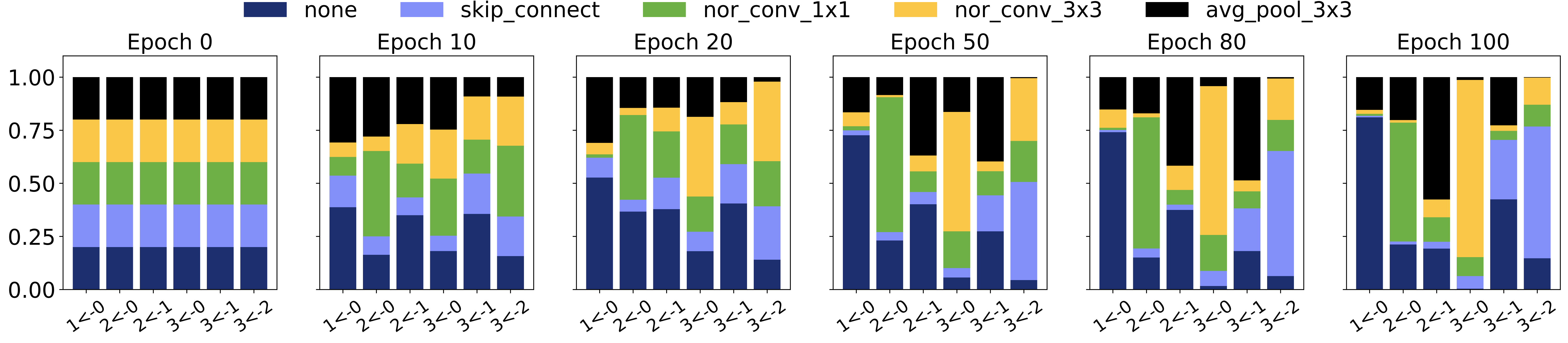}
	}
	\qquad
	\subfloat[]{
		\label{subfig:arch_search_viz_b}
		\includegraphics[width=\linewidth]{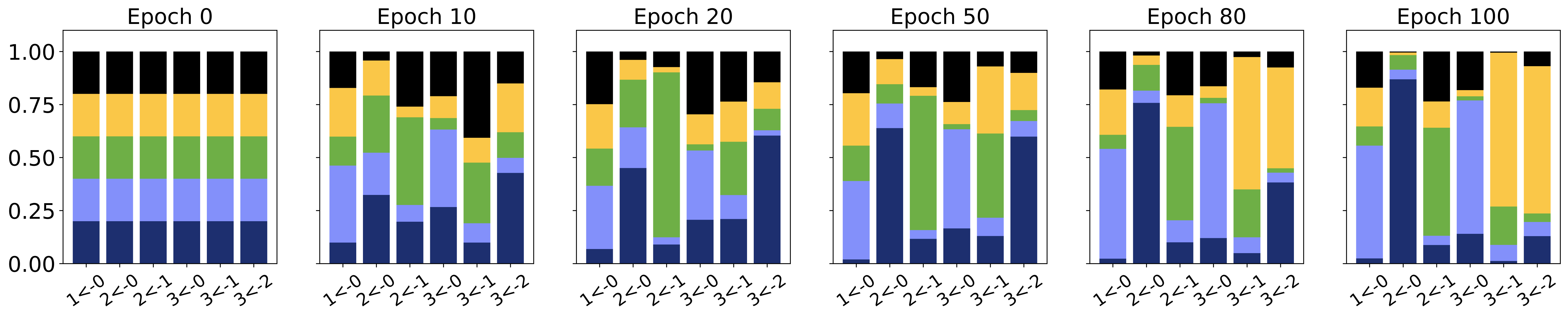}
	}
	\caption{Visualizing the progression of the mean of the normal distribution
		$\mathcal N(\textbf{m}, \textbf{C})$. The colors represent the operations
		in S2 and the width of the color is directly proportional to the weight of
		the operation associated with that color for that specific edge
		(\textit{x-axis}). All the searches were performed on CIFAR-10 for the
		search space S2.
		(a) Architecture search with weight sharing based CMANAS.
		(b) Architecture search with non weight sharing based CMANAS.}
	\label{fig:arch_search_viz}
\end{figure*}

\section{Further Analysis}
In the following sections, we will analyze the different aspects of CMANAS.
\begin{figure}[t]
	\centering
	\begin{center}
		\includegraphics[width=\linewidth]{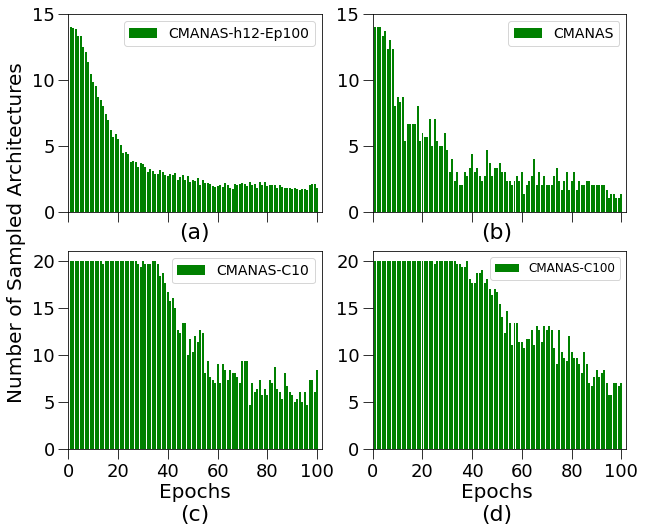}
	\end{center}
	\caption{Number of unique architectures sampled in every epoch averaged
		over all the runs.
		(a)	Non-weight sharing based CMANAS for S2 on CIFAR-10.
		(b) Weight sharing based CMANAS for S2 on CIFAR-10.
		(c) CMANAS for S1 on CIAFR-10.
		(d) CMANAS for S1 on CIAFR-100.
	}
	\label{fig:sample_freq}
\end{figure}
\subsection{Visualizing the Architecture Search}
\label{subsect:viz_arch_search}
\comm{The smaller size of the search space, S2, allow us to visualize the search
process for which we utilize the following techniques:}
For analyzing the search, we visualize the search process by using
the following techniques:
\begin{itemize}
	\item We plotted the number of unique architectures sampled in every epoch for
	both weight sharing based CMANAS and non-weight	sharing based NAS (i.e.
	CMANAS-h12-Ep100) in S2 for CIFAR-10 and the weight sharing based
	CMANAS in S1 for both CIFAR-10 and CIFAR-100 in Fig.~\ref{fig:sample_freq}.
	From the figure, we made the following observations:
	\begin{enumerate}
		\item The number of unique architectures sampled in the beginning part
		of the architecture search was equal to the population size, $N_{pop}$.
		This part could be considered as the \textit{exploration} phase of CMANAS
		wherein the algorithm is exploring the search space by sampling unique
		architectures.
		
		\item As the search progressed, the number of unique sampled architectures
		decreased because of the dominance of some architecture solutions in
		the population. This part could be considered as the \textit{exploitation}
		phase of CMANAS wherein the algorithm keeps on sampling the already
		evaluated good solution.
		
		\item The non-weight sharing based CMANAS (i.e. CMANAS-h12-Ep100)
		converged considerably faster to a better solution than the
		weight sharing based CMANAS in S2. This showed that the trained OSM provided a noisy estimate of the fitness of the architecture.
		
		\item Because of the bigger size of S1 than that of S2, the exploration
		phase of CMANAS was larger in S1 than in S2.
	\end{enumerate}
	
	\item As the architecture parameter, $\alpha$, is modelled using the mean of
	the normal distribution $\mathcal N(\textbf{m}, \textbf{C})$, we can visualize
	the	progression of $\alpha$ by visualizing the mean, $\textbf{m}$, throughout
	the	architecture search epochs. In Fig.~\ref{fig:arch_search_viz}, the mean,
	$\textbf{m}$, of the distribution in an epoch is visualized using a bar
	plot wherein each bar in the plot represents the edge between two nodes in the
	cell (\textit{x-axis}). All the operations in the search space are represented
	by different colors and the weight associated with any operation between two
	nodes (i.e. $\alpha$) is represented by the width of the color associated with
	that operation in the bar associated with that specific edge. From the
	figure, we observed that the search began with equal weights to all the
	operations in all the edges at epoch 0, and as the search progressed, CMANAS
	changes the weights according to the fitness estimation of the architectures 
	in the population. As the search converged to an architecture, CMANAS
	increased the weights of the operations of this architecture in the OSM.
	Fig.~\ref{fig:arch_search_viz} shows the search progression for both
	weight sharing based CMANAS and non-weight sharing based CMANAS for S2 on
	CIFAR-10 dataset only. For CIFAR-100 and ImageNet16-120 datasets, please refer
	to the supplementary.
\end{itemize}

\subsection{Observation on Sampled Architectures}
\begin{figure}[t]
	\centering
	\begin{center}
		\includegraphics[width=\linewidth]{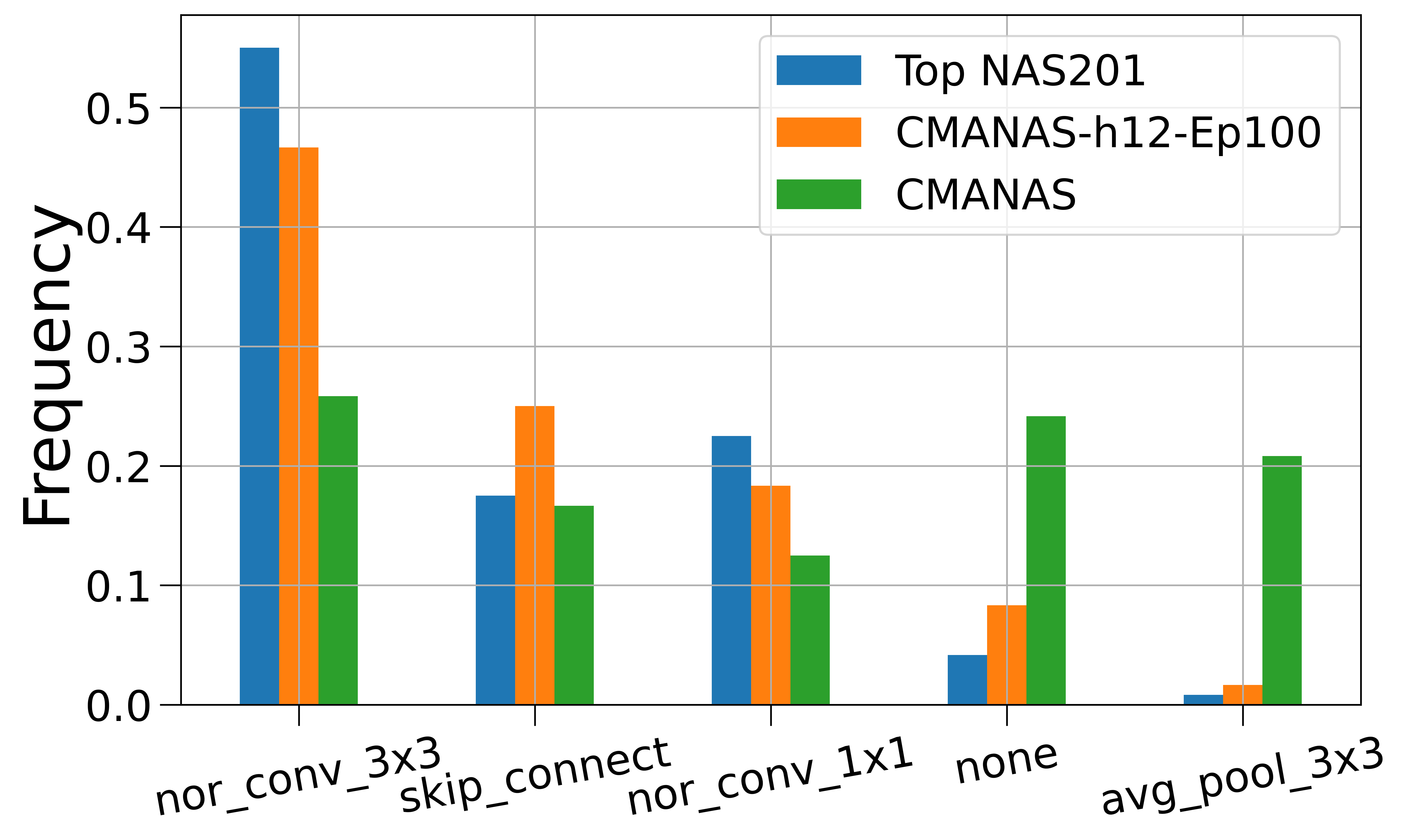}
	\end{center}
	\caption{Frequency of all operations in S2 for top-20 architectures
		in S2, (i.e. NAS-Bench-201 \cite{Dong2020NAS-Bench-201}) labeled as
		"\textit{Top NAS201}", top-20 sampled architectures of non-weight sharing 
		based CMANAS, labeled as "\textit{CMANAS-h12-Ep100}", and top-20 sampled
		architectures of weight sharing based CMANAS, labeled as
		"\textit{CMANAS}".}
	\label{fig:top_op_freq}
\end{figure}
The API provided in the search space, S2, (i.e. NAS-Bench-201
\cite{Dong2020NAS-Bench-201}) allows a faster way to analyze a NAS
algorithm. To discover the patterns associated with the sampled architectures,
we plotted the frequency of all the operations in the search space S2 for the
top-20 sampled architectures (in terms of their estimated fitness) in both the
non-weight sharing based CMANAS (i.e. CMANAS-h12-Ep100) and the weight sharing
based CMANAS in Fig.~\ref{fig:top_op_freq}. They were compared with the frequency
of all the operations in the search space S2 for the top-20 architectures (in
terms of the test accuracy after 200 epochs), for S2, and we observed the
following:
\begin{itemize}
	\item The operation \textit{$3\times3$ convolutions} dominated the top-20
	architectures of S2 which was also seen in the non-weight sharing based
	CMANAS, but the weight sharing based CMANAS showed a marginal dominance.
	
	\item The frequency of operation \textit{$3\times3$ average pooling} was very
	low in the top-20 architectures of S2 and was also seen in the non-weight
	sharing	based CMANAS, but the weight sharing based CMANAS showed higher
	frequency.
\end{itemize}
All these show that the trained OSM provides a noisy estimate of the fitness
of the architecture and with a better estimator, CMANAS can yield better results.

\subsection{Ablation}
\textbf{Comparison with Random Search:} Here, we do not the update the
normal distribution, $\mathcal N(\textbf{m}, \textbf{C})$, after estimating
the fitness of the individual architectures in the population (i.e. no "update
distribution" block in Fig.~\ref{fig:arch_search}). The sampled 
architecture with the best fitness is returned as the searched architecture after
100 epochs. This is essentially a random search and is reported in
Table~\ref{table:c10_s1} for CIFAR-10 as CMANAS-C10rand and for CIFAR-100 as
CMANAS-C100rand. We found that CMANAS-C10rand shows similar results to those
reported in RSPS \cite{li2019random} (random search with parameter sharing). We
also found that the CMANAS took less time than the random search for both datasets
while outperforming the random search in both the datasets.

\textbf{Effectiveness of the AF Table:}
\begin{figure}[t]
	\centering
	\begin{center}
		\includegraphics[width=\linewidth]{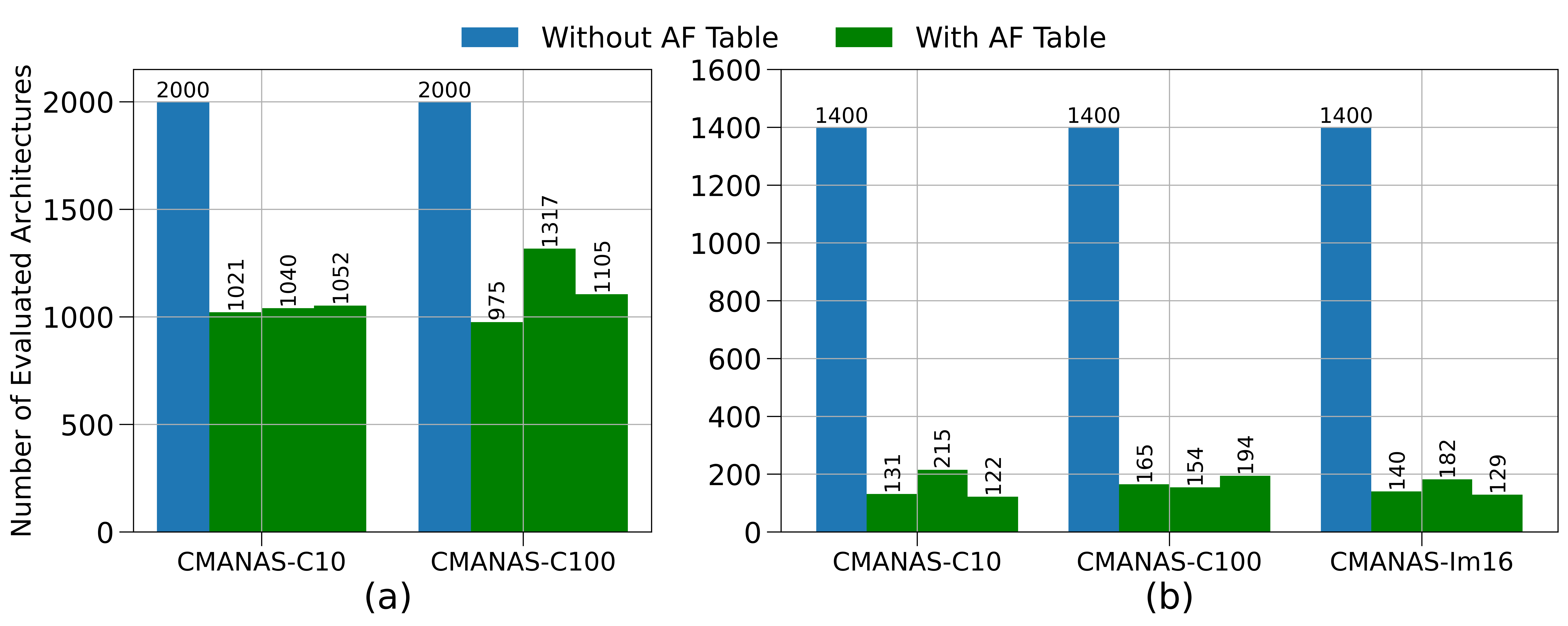}
	\end{center}
	\caption{Total number of architectures evaluated in a single run of weight
		sharing	based CMANAS on
		(a)	S1 for CIFAR-10 and CIFAR-100,
		(b) S2 for CIFAR-10, CIFAR-100 and ImageNet16-120.
	}
	\label{fig:table_stat}
\end{figure}
To illustrate the effectiveness of the AF table used during the evaluation of the
fitness of the architecture, we plotted the total number of architectures
evaluated during the CMANAS search with and without AF table in
Fig~\ref{fig:table_stat}, and made the following observations:
\begin{itemize}
	\item On average, the use of the AF table reduces the architecture evaluation
	by half for S1 and by one-tenth for S2. This results in reducing the search
	time for CMANAS in both the search spaces on all the datasets and has been
	summarized in Table~\ref{table:search_time}.
	
	\item Because of the smaller size of S2 than that of S1, CMANAS requires
	smaller number of architecture evaluations in order to converge to an
	optimal solution in S2 than in S1.
	
	\item The speedup is much bigger for CIFAR-10 as compared to the other
	datasets in both S1 and S2 because of the bigger size of the validation data
	used for CIFAR-10 (Section~\ref{subsect:data_settings}), which results in
	longer fitness evaluation time.
\end{itemize}

\begin{table}[t]
	\caption{Search time of CMANAS with and without the AF table in both S1 and
		S2.	
	}
	\label{table:search_time}
	\centering
	\scalebox{0.77}{%
	\begin{tabular}{|c|cc|ccc|}
		\hline
		& \multicolumn{2}{c|}{\bf{S1}} & \multicolumn{3}{c|}{\bf{S2}}\\
		\hline
		& \multicolumn{2}{c|}{Search time in GPU days} & \multicolumn{3}{c|}{Search time in seconds}\\
		\hline
		&\it{CIFAR-10}& \it{CIFAR-100} & \it{CIFAR-10} &\it{CIFAR-100} &
		\it{ImageNet16-120} \\
		\hline
		Without AF table & $0.66$&$0.67$&$21807$&$27648$&$14220$\\		
		\hline
		With AF table & $0.45$&$0.62$&$13824$&$25992$&$13536$\\		
		\hline
		
	\end{tabular}
	}
\end{table}

\section{Conclusion and Future Work}
The goal of this paper was to develop a framework for applying the Covariance
Matrix Adaptation Evolution Strategy to the NAS problem while using significantly
less computational time than the previous evolution based NAS methods.
This was achieved by using a trained one shot model (OSM) for evaluating the
architectures in the population. The OSM shares weights among all the
architectures in the search space, which allows us to skip training each
individual architecture from sratch for its fitness evaluation. This leads to
reduced search time. We applied CMANAS to two different search spaces
to show its effectiveness in generalizing to any cell-based search space, i.e.
search space agnostic. Experimentally, CMANAS reduced the search time of evolution
based architecture search significantly by 10x to 1000x, while achieving better
results on CIFAR-10, CIFAR-100 and ImageNet datasets than the previous evolutionary
algorithms. CMANAS also solves the overfitting problem present in the
gradinet-based NAS method because of it stochastic nature. We also analyzed the
search process and found that the first part of the architecture search in the
CMANAS acts as the exploration phase, wherein it explores the search space, and
the later part acts as the exploitation phase, wherein the search converges to an
architecture.

A possible future direction to improve the performance of the algorithm is to use
a better fitness estimator as we found that a better estimator would allow CMANAS
to achieve its full potential.


%

\comm{
\appendices
\section{Proof of the First Zonklar Equation}
Appendix one text goes here.

\section{}
Appendix two text goes here.
}

\section*{Acknowledgment}
This work was supported in part by the Ministry of Science and Technology of
Taiwan (MOST 108-2221-E-009-067-MY3 and MOST 110-2634-F-009-018-). Furthermore, we
are grateful to the National Center for High-performance Computing for computer
time and facilities.

\ifCLASSOPTIONcaptionsoff
  \newpage
\fi



\bibliographystyle{IEEEtran}
\bibliography{IEEEabrv, main}
%


%
\comm{\begin{IEEEbiography}[{\includegraphics[width=1in,height=1.25in,clip,keepaspectratio]{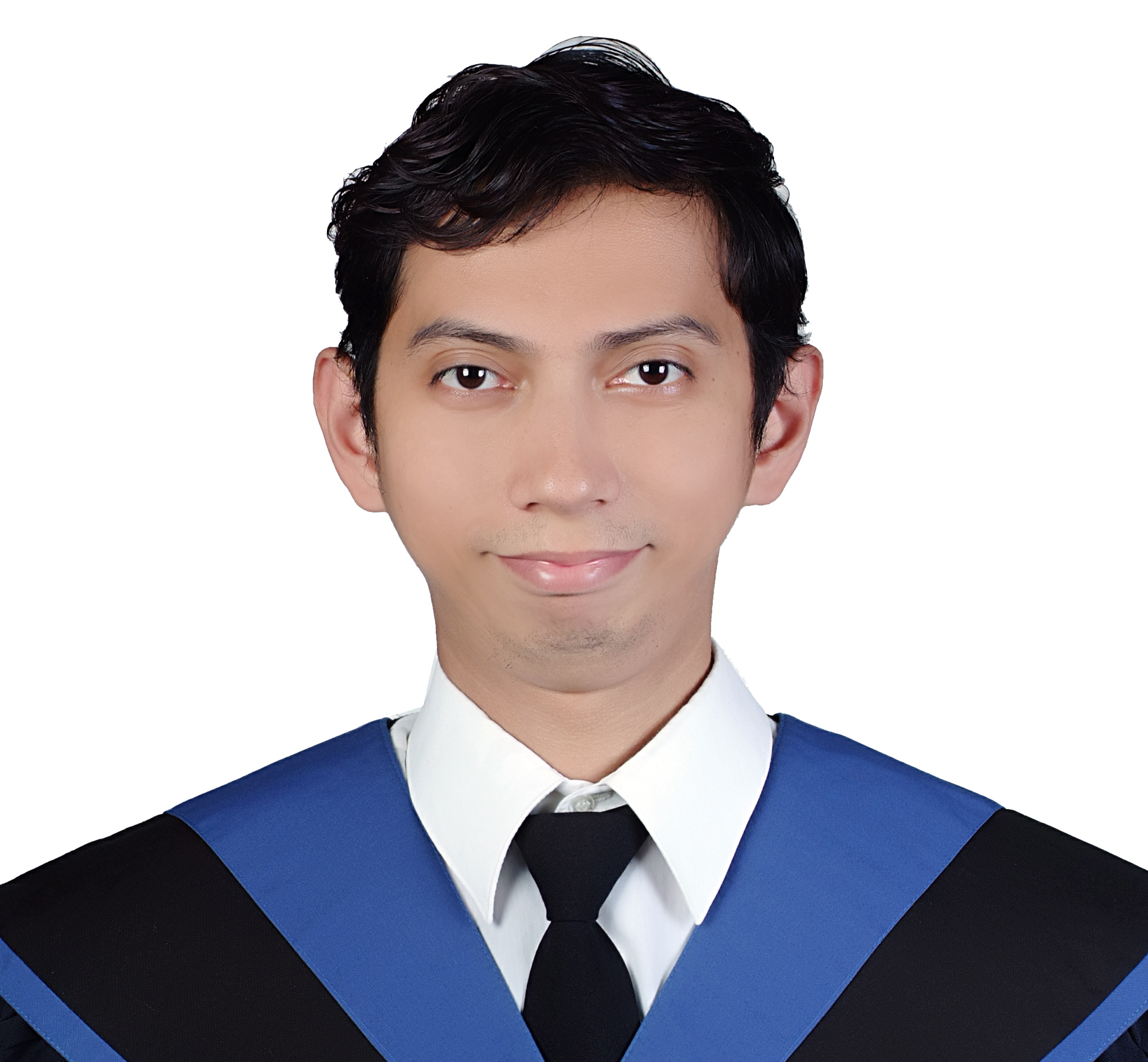}}]{Nilotpal Sinha} received the BTech (Bachelor of Technology) degree
	in Electrical Engineering from National	Institute of Technology, Silchar, India in 2013. He received the MSc degree in Electrical Engineering
	from National Cheng Kung University, Tainan, Taiwan in 2017. He is currently
	a PhD candidate in the CoVis Lab of Department of Computer Science at National
	Yang Ming Chiao Tung University, Hsinchu, Taiwan. His research interests
	include	signal processing, machine	learning and application of evolutionary
	algorithms.
\end{IEEEbiography}

\begin{IEEEbiography}[{\includegraphics[width=1in,height=1.25in,clip,keepaspectratio]{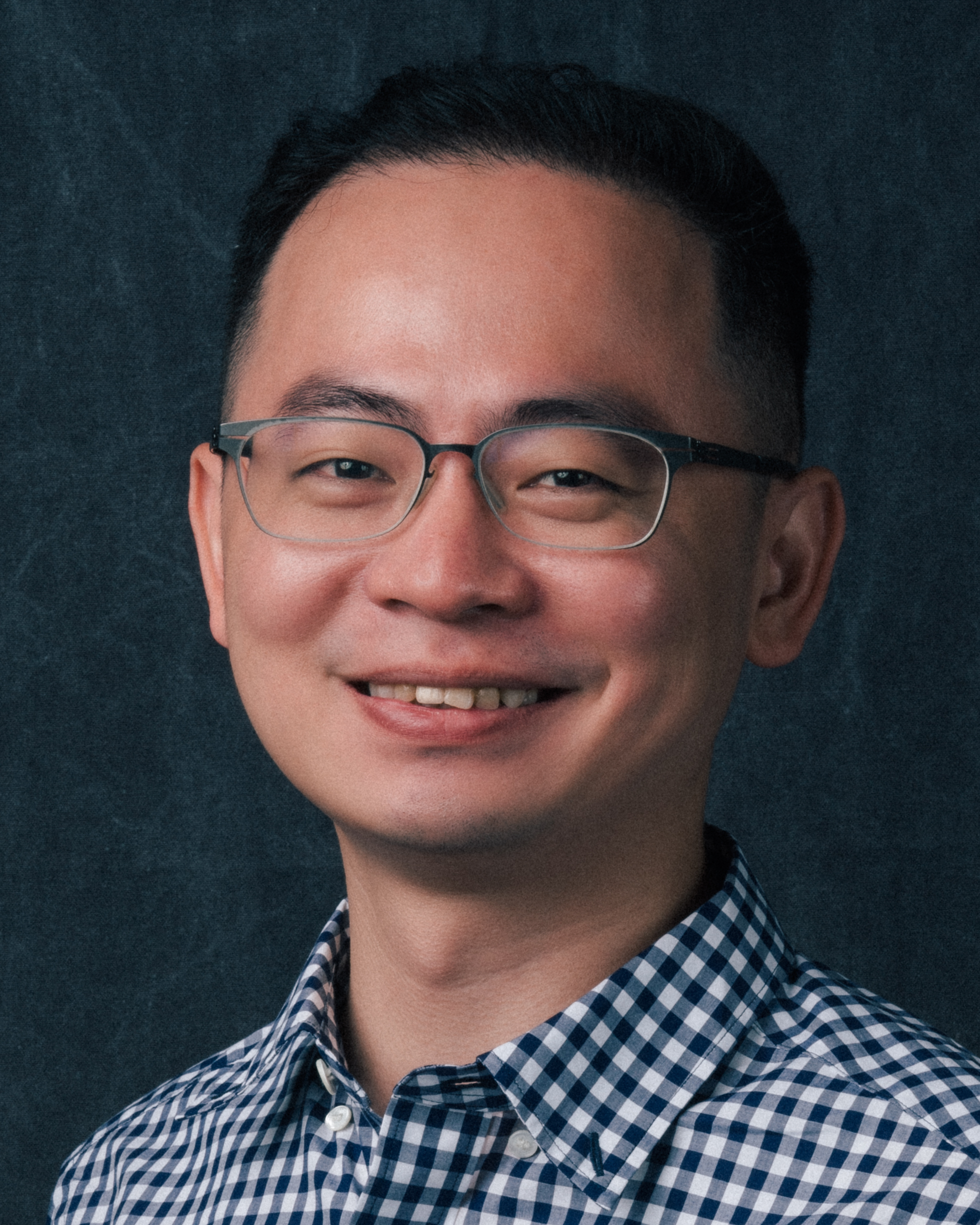}}]{Kuan-Wen Chen} received the B.S. degree in computer and
	information science	from National Chiao Tung University, Hsinchu, Taiwan, in
	2004 and the Ph.D. degree in computer science and information engineering from
	National Taiwan	University, Taipei, Taiwan, in 2011. He is currently an
	assistant professor	with the Department of Computer Science, National Yang
	Ming Chiao Tung	University. From 2012 to 2014, he was a postdoctoral
	researcher with National Taiwan University, where he was an assistant research
	fellow in the Intel-NTU	Connected Context Computing Center from 2014 to 2015.
	His research interests include computer vision, pattern recognition and
	multimedia.
\end{IEEEbiography}

\begin{IEEEbiographynophoto}{John Doe}
Biography text here.
\end{IEEEbiographynophoto}


\begin{IEEEbiographynophoto}{Jane Doe}
Biography text here.
\end{IEEEbiographynophoto}
}




\end{document}